\documentclass[lettersize,journal]{IEEEtran}
\usepackage{amsmath,amsfonts}
\usepackage{algorithmic}
\usepackage{algorithm}
\usepackage{array}
\usepackage{textcomp}
\usepackage{stfloats}
\usepackage{url}
\usepackage{verbatim}
\usepackage{graphicx}
\usepackage{cite}
\usepackage{bm}
\usepackage{subfigure}
\usepackage{multirow}
\usepackage{caption}
\usepackage{cite}
\usepackage{orcidlink}
\usepackage[skip=0pt]{caption}

\begin{document}

\title{GS-GVINS: A Tightly-integrated GNSS-Visual-Inertial\\ Navigation System Augmented by 3D Gaussian Splatting}

\author{Zelin Zhou~\raisebox{2pt}{\orcidlink{0009-0008-3322-7427}}, Saurav Uprety~\raisebox{2pt}{\orcidlink{0009-0007-4694-4268}}, Shichuang Nie~\raisebox{2pt}{\orcidlink{0009-0000-5473-3830}}, and Hongzhou Yang~\raisebox{2pt}{\orcidlink{0000-0002-7579-7582}}, ~\IEEEmembership{Senior Member,~IEEE}
\thanks{\it{(Corresponding author: Hongzhou Yang)}}

\thanks{The authors are with the Department of Geomatics Engineering, Shulich School of Engineering, University of Calgary, Alberta, Canada. (e-mail: zelin.zhou1@ucalgary.ca; saurav.uprety1@ucalgary.ca; sunnie.nie@ucalgary.ca; honyang@ucalgary.ca)}

\thanks{This work has been submitted to the IEEE for possible publication. Copyright may be transferred without notice, after which this version may no longer be accessible.}
}

\markboth{IEEE Transactions on Instrumentation and Measurement. PREPRINT VERSION February~2025}%
{Shell \MakeLowercase{\textit{et al.}}:  Article for IEEE TIM }


\maketitle

\begin{abstract}
Accurate navigation is critical for autonomous vehicles in today’s diverse traffic environments. In recent years, the integration of Global Satellite Navigation System (GNSS), Inertial Navigation System (INS), and camera has demonstrated significant robustness and high accuracy for navigation in complex driving environments. Most integrated systems rely on feature-tracking based visual odometry, which suffers from the problem of feature sparsity, high dynamics, significant illumination changes, outlier sensitivity, etc. Recently, the emergence of 3D Gaussian Splatting (3DGS) has drawn significant attention in the area of 3D map reconstruction and visual SLAM. While extensive research has explored 3DGS for indoor trajectory tracking using visual sensor alone or in combination with Light Detection and Ranging (LiDAR) and Inertial Measurement Unit (IMU), its integration with GNSS for large-scale outdoor navigation remains underexplored. To address these concerns, we proposed GS-GVINS: a tightly-integrated GNSS-Visual-Inertial Navigation System augmented by 3DGS. This system leverages 3D Gaussian as a continuous differentiable scene representation in large-scale outdoor environments, enhancing navigation performance through the constructed 3D Gaussian map. Notably, GS-GVINS is the first GNSS-Visual-Inertial navigation application that directly utilizes the analytical jacobians of SE3 camera pose with respect to 3D Gaussians. To maintain the quality of 3DGS rendering in extreme dynamic states, we introduce a motion-aware 3D Gaussian pruning mechanism, updating the map based on relative pose translation and the accumulated opacity along the camera ray. For validation, we test our system under different driving environments: open-sky, sub-urban, and urban. Both self-collected and public datasets are used for evaluation. The results demonstrate the effectiveness of GS-GVINS in enhancing navigation accuracy across diverse driving environments.
\end{abstract}


\section{Introduction}
\IEEEPARstart{N}{avigation} is a critical component of modern autonomous vehicles and robotics. Continuous, high-accuracy navigation is essential for the optimal performance of advanced driving automation applications. The Global Satellite Navigation System (GNSS) can deliver drift-free centimeter-level global positioning in open-sky environments through real-time kinematics (RTK) positioning mode \cite{ref1}. However, GNSS performance can degrade significantly in scenarios involving non-line-of-sight (NLOS) signals or multipath interference. While Inertial Navigation System (INS) and camera-based visual odometry provide continuous, real-time pose estimation in the local frame, their navigation solutions suffer from error accumulation. To leverage the complementary advantages of each sensor, the GNSS-Visual-Inertial Navigation System (GVINS) \cite{ref2}\cite{ref3}\cite{ref4} has been proposed to enhance localization availability, state estimation accuracy and robustness for navigation in complex environments.

Monocular visual odometry in GVINS leverages visual data and epipolar geometry to estimate camera pose using only a low-cost camera. Most GVINS implementations utilize point-feature tracking methods to introduce additional visual constraints into the batch optimization process for state estimation \cite{ref2}\cite{ref3}\cite{ref4}\cite{ref5}\cite{ref6}. Feature detection plays a crucial role in this process, where corner detectors such as Moravec \cite{ref7}, Harris \cite{ref8}, Forstner \cite{ref9}, Shi-Tomasi \cite{ref10}, FAST \cite{ref11} and ORB \cite{ref12} are commonly employed to extract features. Subsequently, detected features are tracked across sequential frames using Lucas-Kanade (LK) optical flow method \cite{ref13}. Both the 3D positions of the tracked features and the camera poses are jointly optimized in a windowed bundle adjustment framework \cite{ref14}. In this approach, the state estimation problem is formulated to minimize the re-projection errors of the tracked 3D landmarks. Despite its robustness in excluding noisy visual measurement, its performance is highly sensitive to the choice of feature detector and matching thresholds. False matching are more likely in challenging lighting and imaging conditions, such as under direct sunlight, in shadows, or when dealing with image scale variance and blurry image. Additionally, low-texture surfaces can lead to feature sparsity, further increasing the estimation error in visual navigation. Alternatively, Direct Methods leverage the gradient magnitude and direction of image pixel intensities to bypass these limitations \cite{ref15}\cite{ref16}\cite{ref17}. With the exploitation of possibly all information in the image, Direct Visual Odometry (DVO) has demonstrated superior performance in scenarios with low-texture scenes \cite{ref18}, challenging lighting conditions \cite{ref19}, and situations involving camera defocus and motion blur \cite{ref20}, by minimizing photometric errors. 

3D Gaussians have recently emerged as a state-of-the-art (SOTA) scene representation for 3D reconstruction and visual SLAM \cite{ref21}\cite{ref22}. The 3D Gaussian Splatting (3DGS) technique employs explicit and differentiable Gaussian ellipsoids as the sole scene representations, which can be rapidly rasterized into images. By leveraging an efficient tile-based CUDA rasterization algorithm \cite{ref23} and the computational power of modern graphical processing units (GPUs), 3DGS can achieve rendering speeds of up to 200 frame-per-second at 1080p resolution \cite{ref24}. The combination of fast rendering and high-quality novel-view synthesis enables 3DGS-based SLAM to perform accurate real-time camera tracking. Extensive research of 3DGS has been conducted for indoor mapping and trajectory estimation, using either only a camera \cite{ref22}\cite{ref24}\cite{ref25} or a camera integrated with LiDAR or IMU \cite{ref26}\cite{ref27}\cite{ref28}. However, the integration of 3DGS with GNSS for navigation in large-scale, complex outdoor environments remains a research gap.

To address the aforementioned concerns, we present {\bf{GS-GVINS}}, a tightly-integrated GNSS-Visual-Inertial Navigation System augmented by 3DGS. Our system exploits the raw measurement error sources of 3DGS and integrates seamlessly with GNSS RTK tight integration factors, INS factors, feature-tracking based visual factors, motion constraints, and outlier rejection algorithms within a non-linear factor graph optimization (FGO) framework. Six degree-of-freedom (DOF) transformations between the local and global frames are jointly optimized using a sliding-window approach. To account for the fluctuation in 3DGS rendering quality during the frontend tracking phase, we introduce a weighting scheme for 3DGS factors based on the backend mapping loss of keyframes. Furthermore, to ensure a consistent and more stable Gaussian map when the vehicle approaches extreme dynamic states, we propose a Gaussian map pruning mechanism guided by the relative pose translation and ray's accumulated opacity. The key innovations and contributions of this paper are highlighted as follows:

\begin{itemize}
    \item A monocular 3DGS-augmented GNSS-Visual-Inertial Navigation System based on non-linear factor graph optimization is proposed. This is the first application which applies analytical Jacobians of \textit{\textbf{SE}}(3) camera pose with respect to 3D Gaussians within a GVINS framework.  
    \item A weighting scheme for 3DGS factors within the graph optimization process is proposed, leveraging the L1 photometric loss from 3DGS keyframe mapping. This ensures that the importance of image rendering performance for camera tracking can be effectively managed based on the quality of current Gaussian map.    
    \item A motion-aware 3D Gaussian pruning mechanism is proposed to remove unstable Gaussians from the map when the vehicle approaches extreme dynamic states. This ensures the quality of the 3D Gaussian map is maintained during near-stationary motion or rapid motion variations, providing more accurate observations for camera tracking.  
\end{itemize}

\section{Related Works}
\subsection{GNSS-Visual-Inertial Tight Integration}
To achieve a consistently accurate and reliable navigation solution in complex real-world scenarios, multi-sensor fusion is widely recognized as the most effective approach. The tightly coupled integration of GNSS, visual, and inertial data enables the joint optimization of estimated parameters by fusing raw measurements and leveraging all available error sources from each sensor.

In recent years, multiple GVINS frameworks have been developed and published. \cite{ref29} represents the first attempt to integrate raw GNSS measurements of pseudorange and Doppler shift into an optimization-based visual-inertial SLAM system. It demonstrates superior performance compared to existing visual-inertial SLAM and GNSS Single-Point-Positioning (SPP), especially in GNSS-denied environments. GVINS \cite{ref3} introduces a coarse-to-fine initialization technique that accurately establishes the real-time transformation between global measurements and local states, significantly reducing the GNSS state initialization time. Extensive evaluations validate its accuracy and robustness compared to other SOTA visual-inertial-odometry (VIO) systems. However, GVINS supports only the SPP algorithm for GNSS factors, limiting its ability to fully exploit the potential of GNSS positioning. To better utilize the advantages of INS, IC-GVINS \cite{ref4} augments visual feature tracking and landmark triangulation by incorporating precise INS information. This integration significantly improves GVINS performance in high-dynamic conditions and complex environments. However, IC-GVINS only performs GNSS solution-level integration with raw inertial and visual measurements in a batch optimization framework. To fully harness the capabilities of GNSS, GICI-LIB \cite{ref2} incorporates nearly all GNSS measurement error sources in its sensor fusion. Its RTK-RRR processing mode delivers superior pose estimation performance compared to other SOTA GNSS-Visual-Inertial navigation systems, particularly in challenging environments.

Despite these innovations, the primary focus of advancements has been on GNSS and INS. However, further improvement of GNSS or INS performance is increasingly challenging and presents a bottleneck. Meanwhile, visual sensors remain low-cost alternatives, but visual-based navigation still heavily relies on feature tracking and the minimization of reprojection errors. These areas require further attention to enhance overall navigation capabilities in GVINS. With the rapid evolution of graphical computing resources, more sophisticated visual representations and computer graphics techniques are being developed and can now be efficiently applied to address a variety of engineering challenges. These advancements open new opportunities to push the boundaries of visual navigation.

\subsection{3DGS in SLAM}

Traditional SLAM systems rely on points \cite{ref30}, surfels \cite{ref31} or voxel grids \cite{ref32} as scene representations, enabling direct and fast computations. However, these methods often struggle to achieve high-fidelity mapping due to their fixed spatial resolution and the lack of correlation among the 3D primitives. Neural-based SLAM \cite{ref33}\cite{ref34}\cite{ref35} offers improved mapping quality but suffers from computationally expensive training processes, making it unsuitable for many real-time applications. In contrast, 3DGS has demonstrated superiority in efficient, real-time, and high-resolution image rendering. This is achieved by using expressive anisotropic 3D Gaussians as scene representations and leveraging tile-based rasterization techniques. As a result, 3DGS-based SLAM systems outperform traditional SLAM methods by enabling rapid photo-realistic rendering, and achieve superior performance and efficiency in both mapping and tracking. GS-SLAM \cite{ref22} is the first to incorporate real-time 3DGS into the SLAM pipeline using RGB-D rendering. It integrates an adaptive expansion strategy to add or remove noisy 3D Gaussians, which enhances the mapping performance. Additionally, GS-SLAM introduces an effective coarse-to-fine method for selecting reliable 3D Gaussians, enhancing camera pose optimization. MonoGS \cite{ref24} represents the first application of 3DGS in real-time monocular SLAM. It replaces the offline Structure-from-Motion (SfM) process used in the original 3DGS algorithm with direct optimization against 3D Gaussians, taking advantage of their wide convergence basin for robust camera tracking. MonoGS also introduces a geometric verification and regularization technique to address ambiguities in incremental 3D dense reconstruction. SplaTAM \cite{ref25} proposes a single RGB-D camera SLAM framework that utilizes a silhouette mask to capture the presence of scene density. Both online tracking and mapping evaluations show that SplaTAM outperforms existing methods. Despite these advancements, 3DGS-based SLAM faces challenges, including error accumulation during pose tracking, scale ambiguity in monocular 3DGS SLAM, and degradation of pose tracking caused by inconsistent or erroneous 3D Gaussian maps. Our navigation system takes advantage of the tight integration of 3DGS with GNSS and INS to address these challenges.

\section{System overview}

\setlength{\textfloatsep}{15pt}

\begin{figure*}[t]
\centering
\includegraphics[width=1.0\textwidth, height=10cm]{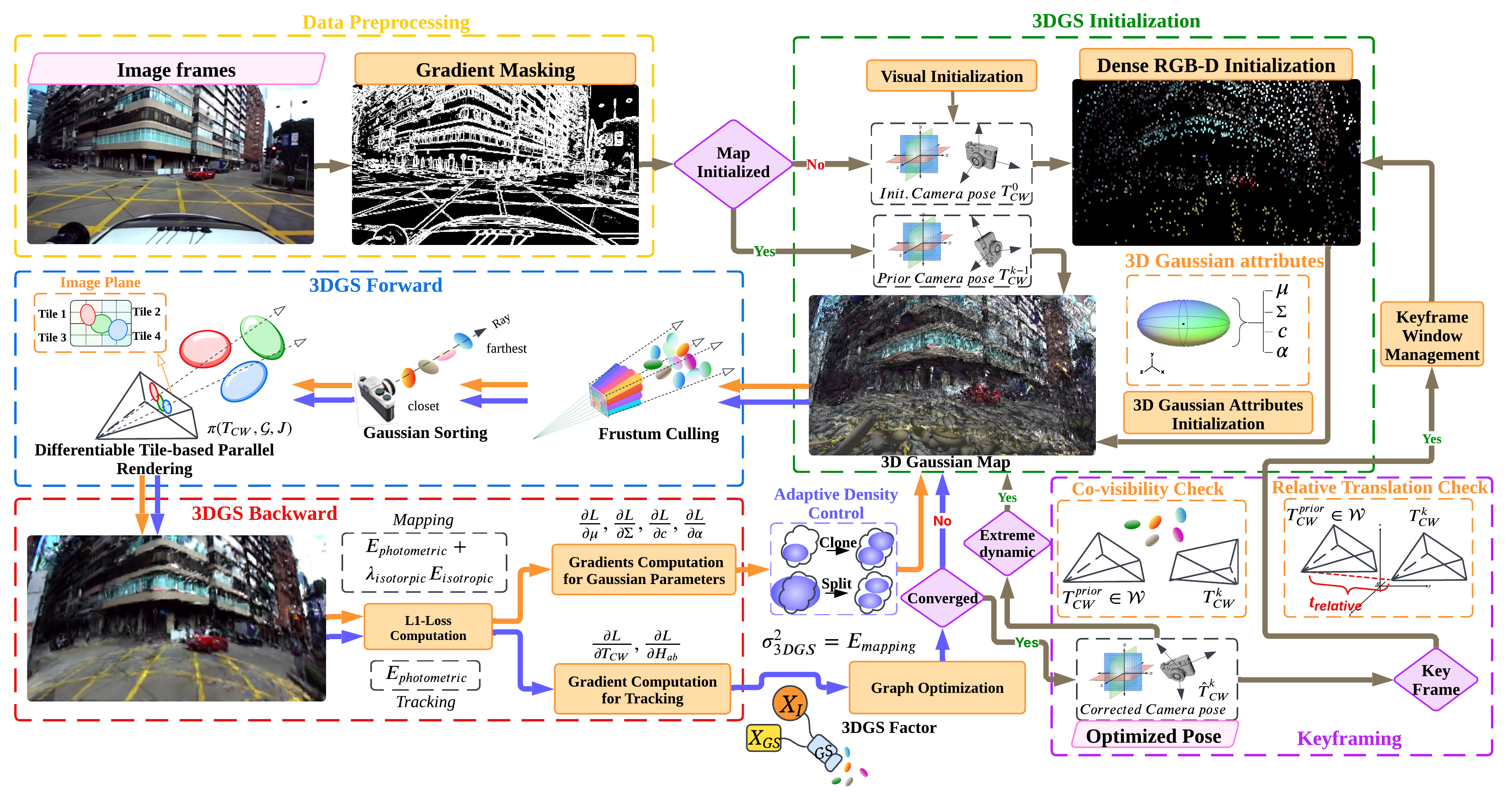}
\caption{{\bf{Overview of the 3DGS module:}} The module performs 3D Gaussian optimization (orange arrows) and pose optimization (purple arrows), each operating in its own thread and intercommunicating.}
\label{fig_1}
\end{figure*}

GS-GVINS adopts most of its system architecture, including the initialization algorithm, marginalization, and FGO pipeline (specifically for the tight integration of GNSS RTK, INS, and feature-based visual factors) from \cite{ref2}. Additionally, we introduce a 3DGS module to support 3DGS factors. The system comprises I/O hardware controllers, data decoders/encoders, data streamers, and estimation processors. Each component operates in its own thread, enabling real-time, pseudo-real-time, and post-processing configurations. As the implementations of other factors are thoroughly documented in the system manual of \cite{ref2}, this paper focuses on the 3DGS factor and its implementation.  

Figure 1 illustrates the architecture of the 3DGS module, which can be modularized into data preprocessing, 3DGS initialization, 3DGS forward, backward and keyframing. Each incoming RGB image in the sequence is masked using binary gradients (based on a preset threshold) to exclude weak edges. If the 3D Gaussian map has not been initialized, the first frame is used for initialization. With the initial estimated pose $\pmb{\mathit{T}}_{CW}^0$ obtained from GNSS/IMU and visual initialization, a dense RGB-D point cloud is generated to establish the initial colors and 3D positions (coarse depths are initialized and will be optimized) of the Gaussian in the world frame. After initializing each Gaussian’s scaling, rotation, and opacity, frustum culling removes any Gaussians outside the camera's field of view. Overlapping Gaussians along each pixel’s line of sight are then sorted by depth. The 3D Gaussian ellipsoids are projected onto the 2D image plane using the estimated camera pose and projection matrix, and the tile-based parallel CUDA rendering synthesizes the final pixel colors. An $L_1$ photometric loss is computed by comparing the rendered image with the ground-truth image. During back-propagation, the Jacobians of the Gaussian attributes and the camera pose are evaluated. 

For mapping, the positional gradients enable adaptive density control of the 3D Gaussians, allowing Gaussians to be split or cloned to better represent fine geometric details. A small keyframe window is maintained, adding new keyframes for novel views and removing outdated ones. New Gaussians are also inserted to capture additional scene details, while redundant Gaussians are pruned. The 3D Gaussian map within the current window is then jointly optimized.

With the 3D map constructed in this manner, pose tracking is performed by re-rendering the current viewpoint, comparing it against the incoming frame, and minimizing the photometric loss through camera pose optimization.  

\begin{figure}[!t]
\centering
\hspace{-2cm} 
\includegraphics[width=3.4in]{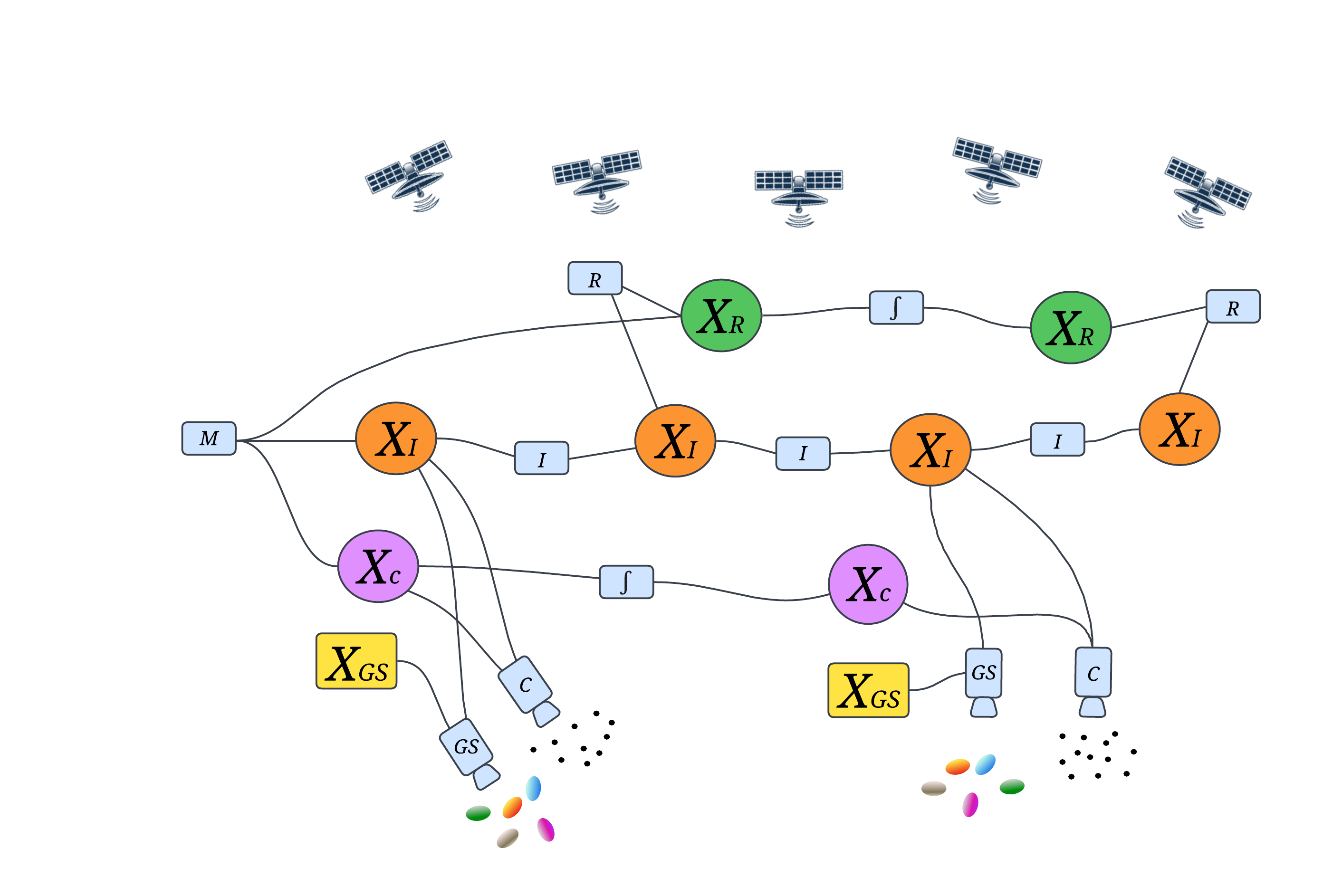}
\caption{Factor graph structure in GS-GVINS.}
\label{fig_1}
\end{figure}

\section{Methodology}

The key innovations of this paper are demonstrated by focusing on the elaboration of the 3DGS algorithm and its integration into sensor fusion.

In this section, we first introduce the 3DGS-augmented navigation. Next, we elaborate on the weighting scheme for the 3DGS factor based on the 3DGS mapping loss. Finally, we present the motion-aware Gaussian map pruning algorithm.

\subsection{3DGS-augmented Navigation}

In GS-GVINS, a factor graph representation is employed to describe the non-linear Least-Squares (LSQ) problem, where the estimated parameters and measurements are represented as nodes, and the residuals are represented as edges. Figure 2 illustrates the factor graph of the LSQ problem in GS-GVINS, where the Gaussian states are represented by yellow rectangles as they are optimized independently from the FGO. The solution of the LSQ problem can be defined by (1).

\begin{equation}
\begin{aligned}
    \bm{\hat{\chi}} = \operatorname*{arg\,min}_{\bm{\chi}} \Bigg\{
    &\left\| \bm{z}_p - H_p \bm{\chi} \right\|^2 
    + \left\| \bm{z}_r - h_r(\bm{\chi_r}, \bm{\chi_I}) \right\|^2 \\
    &+ \left\| \bm{z}_I - h_I(\bm{\chi_I}) \right\|^2 
    + \left\| \bm{z}_c - h_c(\bm{\chi_c}, \bm{\chi_I}) \right\|^2 \\
    &+ \left\| \bm{z}_{GS} - h_{GS}(\bm{\chi_I}) \right\|^2
    \Bigg\}
\end{aligned}
\end{equation}

\begin{equation}
    \bm{\chi}_r=[^{B}t^{T}_{r}, ^{W_{G}}p^{T}, ^{W_{G}}v^{T}, df_{r}, N^{s}_{rr_{b},i}]^{T}
\end{equation}

\begin{equation}
    \bm{\chi}_I=[^Wp^T, {q^W_{B}}^T, ^Wv^T, b_a,b_g]^{T}
\end{equation}

\begin{equation}
    \bm{\chi}_c=[^Bt_c, q^B_C,^Wp_l]^{T}
\end{equation}

The state vector is defined as $\bm{\chi}=[\bm{\chi}_r,\bm{\chi}_I,\bm{\chi}_c]$, where the subscripts $r$, $I$, $c$ and $GS$ correspond to GNSS receiver, INS, camera, and 3DGS, respectively. Their full elements are shown in (2), (3) and (4). $^{B}t^{T}_{r}$ represents the GNSS antenna extrinsic parameters, $^{W_{G}}p^{T}$ and $^{W_{G}}v^{T}$ denote the position and velocity of the vehicle body in Earth-Centered Earth-Fixed (ECEF) frame, $df_{r}$ represents the frequencies for each satellite system, $N^{s}_{rr_{b},i}$ refers to single-difference carrier phase ambiguities. For the INS states, $^Wp^T$, ${q^W_{B}}^T$ and $^Wv^T$ represent the position, orientation and velocity of the vehicle body in East-North-Up (ENU) world frame. Its origin is defined as the first estimated GNSS position during system initialization. $b_a$ and $b_g$ are the biases of accelerometer and gyroscope. For the feature-based camera states, $^Bt_c$ and $q^B_C$ represent the camera's extrinsic parameters, and $^Wp_l$ refers to the 3D positions of landmarks in ENU frame. $\bm{z}$ denotes the measurements and $h$ are the non-linear measurement models.  $\bm{z}_p$ and $H_p$ represent the pseudo-measurements and linearized Jacobian of the prior information, which are computed during marginalization. Note that $\bm{z}_c$, $h_c$ and $\bm{\chi}_c$ only refer to the information related to the image-feature-based reprojection error factor. In GS-GVINS, 3D Gaussian parameters are optimized independently from the FGO, considering only the INS states in the 3DGS measurement model.

3DGS factor applies the principle of differentiable 3DGS to provide additional information for pose estimation. In our monocular case, the $\mathcal{L}_1$ photometric error can be modeled based on the current Gaussian map and the estimated camera pose, illustrated in (5). 

\begin{equation}
\label{deqn_ex1a}
E_{photometric}= \left\| I(\mathcal{G},\bm{T}_{CW}) - \bar{I} \right\|_1
\end{equation}

where $I(\mathcal{G},\bm{T}_{CW})$ represents the rendered image from the 3D Gaussian map $\mathcal{G}$ based on $\bm{T}_{CW}$. And $\bar{I}$ is the ground truth image from observation. In the loss computation, optimized affine brightness parameters are utilized to account for varying exposure conditions. An RGB boundary mask is applied to focus the penalty on edge pixels while disregarding non-edge regions. Additionally, the opacity parameter derived from rendering is incorporated to penalize pixels with low-opacity values, ensuring a more robust loss evaluation. We optimize the estimated camera pose by minimizing the photometric error in (5).

In the forward process of 3DGS, the RGB-D point clouds extracted from the image are initialized as 3D anisotropic Gaussians $\mathcal{G}$. In our outdoor monocular setting, depth estimates are initialized with a large integer value, added by noise with high variance. Each Gaussian $\mathcal{G}^i$ is characterized by its optical properties: color $\bm{c}^i$ and opacity $\bm{\alpha}^i$, and its geometric properties: mean (3D position of center) $\bm{\mu}^i_W$ and covariance (ellipsoidal shape) $\bm{\Sigma}^i$, defined in world space.
\begin{equation}
    \mathcal{G}_i(\bm{x})=e^{-\frac{1}{2}(\bm{x})^T\bm{\Sigma}^{-1}_i(\bm{x})}
\end{equation}
The view-dependent radiance described by spherical harmonics (SHs) are omitted in this work for performance efficiency and simplicity. 

After 3D Gaussian initialization, the rasterization iterates over each $\mathcal{G}^i$ and project it from 3D world space $\mathcal{G}(\bm{\mu}_W, \bm{\Sigma}_W)$ into 2D image space $\mathcal{G}'(\bm{\mu}_I, \bm{\Sigma}_I)$ through a projective transformation, illustrated by (7).

\begin{equation}
    \bm{\mu}_I =\pi(\bm{T}_{CW} \cdot \bm{\mu}_W),\quad \bm{\Sigma}_I=\bm{J}\bm{W}\bm{\Sigma}_W\bm{W}^T\bm{J}^T 
\end{equation}
where $\pi$ is the projection matrix and $\bm{T}_{CW}\in SE(3)$ represents the camera pose that transform world frame into camera frame. $\bm{J}$ is the Jacobian of the affine approximate of the projective transformation and $\bm{W}$ is the rotation part of $\bm{T}_{CW}$. Given the 3D position of the center of each $\mathcal{G}^i$, we formulate a list $\mathcal{N}$ that sorts all the overlapping Gaussians in 3D camera-view space with respect to the ray of each pixel, based on their depths. By traversing $\mathcal{N}$ from front to back, the final pixel color in the 2D image space is synthesized through the volumetric rendering along a ray, as shown in (8)
\begin{equation}
    C=\sum_{n=1}^{\mathcal{N}}c_n\alpha_n'\prod_{j=1}^{n-1}(1-\alpha_j')
\end{equation}
\begin{equation}
    \alpha_n'=\alpha_n\times e^{-\frac{1}{2}(\bm{x}'-\bm{\mu}_n')^T\bm{\Sigma}_n'^{-1}(\bm{x}'-\bm{\mu}_n')}
\end{equation}

where $c_n$ is the optimized color of $\mathcal{G}^n \in \mathcal{N}$, $\alpha_n'$ is the opacity value in projected Gaussian space, explained by (9), and $\alpha_j'$ represents the opacity of each previous Gaussian in the list. The per-pixel contribution from the listed Gaussians are decayed in the order of sequence, by the transmittance based on all previous blended Gaussians. In (9), $x'$ and $\mu_n'$ are the coordinates of image pixel and the projected Gaussian 's center. To facilitate the rasterization efficiency, tile-based parallel rendering is applied in CUDA programming architecture. 

In the backward process of 3DGS, the calculated loss value is used to compute the gradients through backward-propagation. 

For mapping, both optical parameters and geometric parameters of 3D Gaussians are optimized through iterative rendering against the training view and first-order Stochastic Gradient Descent (SGD) techniques. Since the Gaussians' attributes do not have correlation with parameters of GNSS or INS, the mapping is conducted independently from the FGO. Keyframes are selected based on the co-visibility of Gaussian map and the relative translation of vehicle. If the intersection of constructed Gaussians between the current frame and the last keyframe drops below a threshold, or if the relative translation since the last keyframe surpass the median depth of the current Gaussian map, the current frame is registered as a new keyframe. A small window of keyframes is maintained to capture the scene in the surrounding environment. A keyframe will be removed from the window if its overlap with the latest keyframe drops below a threshold, along with its associated 3D Gaussians, which will also be removed from the map. Furthermore, we prune redundant Gaussians based on their visibility. In the outdoor monocular case, accurately estimating the 3D positions of many Gaussians remains challenging, and those associated with dynamic objects should be excluded from the map. These Gaussians usually violate multi-view consistency and will quickly vanish during optimization. To refine the 3D Gaussians map during optimization, we follow the same procedure as the original 3DGS work for adaptive density control. We further adopt the isotropic regularization techniques from \cite{ref24} to penalize the Gaussians which are highly elongated along the viewing direction and cause artefacts during rendering, further improve the tracking accuracy.

During tracking, both camera pose and the affine brightness parameters are optimized. To reduce the computational overhead of automatic differentiation, the explicit analytical Jacobian of $\bm{SE}(3)$ camera pose with respect to 3D Gaussians, derived from \cite{ref24}, is integrated into our FGO pipeline. In (7), both $\bm{\mu}_I$ and $\bm{\Sigma}_I$ are differentiable with respect to $\bm{T}_{CW}$. The minimal Jacobians of 6-D world-camera transformation can be derived using Lie algebra as:
\begin{equation}
    \frac{\partial\bm{\mu}_I}{\partial\bm{T}_{CW}}=\frac{\partial\bm{\mu}_I}{\partial\bm{\mu}_C}\frac{\mathcal{D}\bm{\mu}_C}{\mathcal{D}\bm{T}_{CW}}
\end{equation}

\begin{equation}
    \frac{\partial\bm{\Sigma}_I}{\partial\bm{T}_{CW}}=\frac{\partial\bm{\Sigma}_I}{\partial\bm{J}}\frac{\partial\bm{J}}{\partial\bm{\mu}_C}\frac{\mathcal{D}\bm{\mu}_C}{\mathcal{D}\bm{T}_{CW}}+\frac{\partial\bm{\Sigma}_I}{\partial\bm{W}}\frac{\mathcal{D}\bm{W}}{\mathcal{D}\bm{T}_{CW}}
\end{equation}

The derivatives of camera pose lay on the manifold for minimal parametrization, ensuring the dimension of the Jacobians match the exact degrees of freedom of $\bm{SE}(3)$. The partial derivatives with respect to $\bm{T} \in SE(3)$ can be described by (12)

\begin{equation}
    \frac{\mathcal{D}f(\bm{T})}{\mathcal{D}\bm{T}} \overset{\triangle}{=}\lim_{\tau \to 0}\frac{\mathrm{Log}(f(\mathrm{Exp}(\tau)\circ )\circ f (\bm{T})^{-1})}{\tau}
\end{equation}
where $\tau \in \mathit{se}(3)$ represents the 6-D twists on the tangent space of manifold, $\circ$ is a group composition, $\mathrm{Log}$ and $\mathrm{Exp}$ are the logarithmic and exponential mappings between Lie Group and Lie algebra. Therefore, the Jacobians of camera pose in (10) and (11) are derived as:
\begin{equation}
    \frac{\mathcal{D}\bm{\mu}_C}{\mathcal{D}\bm{T}_{CW}}=[\bm{I} 
 \qquad  \bm{\mu}_C^{\times}],\quad \frac{\mathcal{D}\bm{W}}{\mathcal{D}\bm{T}_{CW}}=\begin{bmatrix}
     \bm{0} & -\bm{\mathrm{W}}^{\times}_{:,1}\\
     \bm{0} & -\bm{\mathrm{W}}^{\times}_{:,2} \\
     \bm{0} & -\bm{\mathrm{W}}^{\times}_{:,3}
 \end{bmatrix}
\end{equation}

where $^{\times}$ represents the skew symmetric matrix form, $\bm{\mathrm{W}}^{\times}_{:,i}$ indexes the $i$th column of the rotation matrix.

\subsection{3DGS Mapping Loss based Weighting Scheme}

The 3DGS factor computes analytical Jacobians of camera pose using the $\mathcal{L}_1$ photometric loss by rendering the 3D Gaussian map at viewpoint and compare it with the ground truth image. The quality of rendering strongly depends on the accuracy of the optimized 3D Gaussian map. A Gaussian map optimized with a higher mapping loss introduces more rendering errors in camera tracking, resulting in less reliable pose estimation. 

In the window-based mapping strategy, all keyframes within the window contribute to the map optimization at a given viewpoint. However, in complex large-scale and continuous driving scenarios, it is infeasible to optimize a 3D Gaussian from multiple viewing directions. The optimized 3D Gaussians may appear distorted when rendered from different viewing directions later. This is a typical issue that can degrade tracking performance due to the resulting erroneous rendering.  

To account for the uncertainties from the 3DGS mapping optimization, we introduce a weighting scheme for 3DGS factor based on its mapping loss. The full weighting scheme can be described by (14). 

\begin{equation}
    \bm{\sigma}^{2}_{\mathrm{3DGS},i}=\sum_{\forall{k}\in\mathcal{W}}E^{k}_{photometric}+\lambda_{isotropic}E_{isotropic}
\end{equation}

\begin{equation}
    E_{isotropic}=\sum_{j=1}^{|\mathcal{G}|}\|s_j-\Tilde{s}_j\cdot\bm{1}\|_1
\end{equation}

where $\bm{\sigma}^{2}_{\mathrm{3DGS},i}$ is the variance of 3DGS factor at the $i_{th}$ frame, $\mathcal{W}$ is the current keyframe window, $k$ represents the $k_{th}$ keyframe, $E^{k}_{photometric}$ is the photometric loss computed using (5), $\lambda_{isotropic}$ is the pre-defined isotropic weighting factor, $s_j$ and $\Tilde{s}_j$ are scaling parameters and its mean for the $j_{th}$ Gaussian in the map. This considers the contribution of error sources from all the keyframes in the current window and the isotropic regularization of over-stretched Gaussians.

\subsection{Motion-Aware 3D Gaussian Pruning}

When the vehicle approaches extreme dynamic states, such as near-static conditions or high-dynamic motion, camera tracking performance during non-keyframes can be impacted by noisy rendering. Erroneous renderings are more likely to occur due to positional errors in 3D Gaussians and the predominantly forward-oriented motion of the vehicle. These errors may cause 3D Gaussians to appear misplaced or excessively close to the viewpoint, leading to rendering artifacts such as blurriness, holes, or floating elements, as shown in Figure 8.

The optimization of the 3D Gaussian map relies on all keyframes within the current window. However, the probabilities of keyframe registration decrease significantly when the vehicle undergoes rapid deceleration or comes to a halt. As a result, the current Gaussian map is updated much less frequently, as no modifications occur within the keyframe window. Additionally, due to the inherent scale ambiguity in a monocular setting, the optimized map may appear distorted when viewed from different angles. This distortion becomes more pronounced during high-dynamic motion, where larger displacements between sequential viewpoints introduce relatively more significant rendering artifacts, particularly in forward-oriented motion. As a result, rendering from extreme dynamic viewpoints can lead to high photometric loss during tracking.

To address this issue, we propose a motion-aware 3D Gaussian pruning mechanism. This approach removes redundant 3D Gaussians from the map that could introduce noise in the rendering during extreme motion. The pruning process is guided by the computed relative translation between the consecutive frames, and the accumulated opacity $\alpha$ from the rays that capture the most recently inserted 3D Gaussians (from the latest keyframe). This mechanism is illustrated in (16).

\begin{equation}
\bm{\mathcal{G}}_{static}=
\begin{cases}
    {\mathcal{G}}\backslash\mathcal{G}_{redundant}, & \text{if } t_i < \lambda_{min} \text{ or } t_i > \lambda_{max}, \\
    & \forall i \in [n-9, n], \\
    \mathcal{G}, & \text{otherwise}.
\end{cases}
\end{equation}

\begin{equation}
\mathcal{G}_{redundant} = \bigcup_{k \in \mathcal{R}} \{ \mathcal{G}_{j} \mid \sum_{j \in R_k} \alpha_j \geq 0.5 \}
\end{equation}

where $\mathrm{t}_i$ is the relative translation of the $i_{th}$ frame with respect to its previous frame, and $\lambda_{min}$ and $\lambda_{max}$ are the predefined thresholds of relative translation for low-dynamic and high-dynamic case, respectively; $\mathcal{G}$ denotes the set of 3D Gaussians in the current map, while $\mathcal{G}_{redundant}$ represent the Gaussians to be pruned. $\mathcal{R}$ is the set of rays that intersect the newly inserted 3D Gaussians, with $k$ denoting an individual ray from this set. $R_k$ represents the set of Gaussians along the $k^{th}$ ray that are positioned in front of the newly inserted Gaussians. Finally, $\mathcal{G}_{j}$ refers to each individual 3D Gaussian along the $k^{th}$ ray.

If the displacement between consecutive viewpoints falls below or exceeds predefined thresholds, we infer that the vehicle is undergoing extreme dynamic motion. As described in (16), 3D Gaussians in the current map are pruned by removing those originating from older keyframes and preventing the integration of new 3D Gaussians from recent keyframes. This approach ensures that the latest keyframe, which generally provides a clearer view and a more accurate camera-to-3D Gaussian orientation, is prioritized in extreme motion scenarios.

After sorting the 3D Gaussians along the camera ray, we identify and remove those that cause the accumulated opacity $\alpha$ to reach 0.5 before blending with new Gaussians. This process reduces the density of closer 3D Gaussians in the vehicle's heading direction, preventing unstable Gaussians from accumulating in regions with near-stationary motion or rapid motion variations.

\section{Experiments and Results}

To validate our proposed system, we conduct extensive experiments using both the open-sourced UrbanNav \cite{ref38} dataset and self-collected data, covering a variety of GNSS environments ranging from open-sky to deep urban scenarios. All datasets are recorded in kilometers-scale large outdoor environments with several vehicle turns, stops, and elevation variations. The evaluation focuses on absolute pose error (APE) and compares GS-GVINS results with GICI-LIB and IC-GVINS, which are SOTA open-sourced GVINSs. 

This section is structured as follows: First, we briefly describe the experimental setup for our self-collected data and evaluation. Next, we detail the dataset characteristics and the configurations used in the evaluation. Finally, we present and analyze the results.

\subsection{Experimental Setup}

For our self-collected data, we utilized a multi-sensor integrated platform equipped with hardware-based time synchronization (milliseconds level accuracy) to collect GNSS rover, inertial, and visual raw measurements. The platform consists of a u-blox F9P GNSS receiver, an ICM-20948 MEMS IMU, and two FLIR Blackfly-S RGB cameras. An STM32 Micro-Controller-Unit (MCU) manages system time synchronization using the GPS PPS signal, as well as data streaming and storage. A u-blox patch antenna is connected to the u-blox F9P receiver. The GNSS, inertial and visual data are collected in 1 Hz, 100 Hz and 10 Hz respectively. For groundtruth generation, the system is equipped with a NovAtel SPAN system comprising an OEM7 receiver and an Epson EG320N IMU. The time synchronization algorithm and the full details can be found in \cite{ref39}. GNSS reference station data were collected from a Trimble R9 receiver located on the rooftop of the CCIT building at the University of Calgary. The baseline distance between the reference station and the rover consistently remained below 10 km. All data are stored on a laptop using ROS 2 \cite{ref40}.

\begin{figure}[htbp]
    \centering
    \includegraphics[width=1.0\columnwidth]{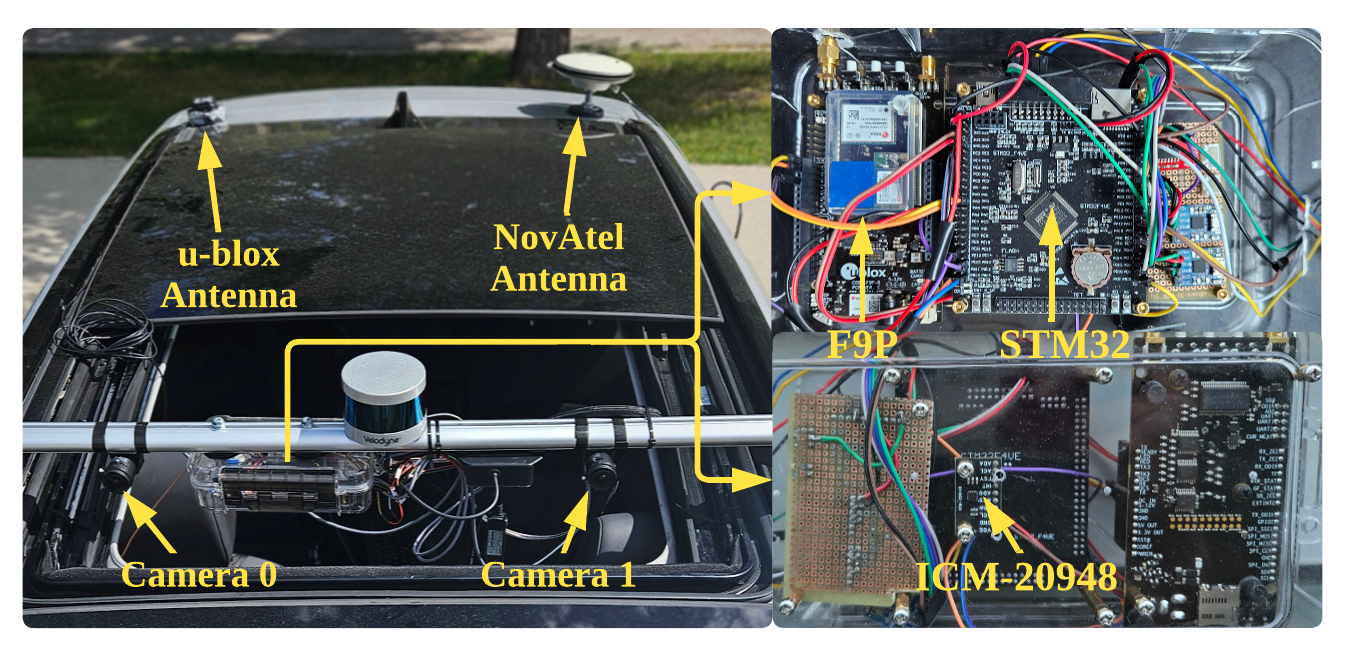}
    \caption{Multi-sensor data collection system mounted on a vehicle (left), with top view (upper-right) and bottom view (lower-right) of the control box.}

\end{figure}

To evaluate our proposed system, we replay the ROS bag data in a pseudo-real time configuration. The data are published via ROS topics and streamed into the GS-GVINS in the order of their message timestamps. The entire system operates within the environment of Advanced Computing Cluster (ARC) at the University of Calgary. During runtime, the system utilizes 4 out of 7 units of an NVIDIA A100-MIG GPU with 40 GB VRAM, 4 cores of an Intel(R) Xeon(R) Silver 4316 CPU running at 2.30 GHz, and 70 GB of RAM.

\subsection{Dataset Characteristics and Configuration}

Figure 4 provides views of scenes from each validation dataset. For consistency, we categorize the environments based on urbanization levels into three types: open-sky, sub-urban, and urban. Two self-collected datasets are used to evaluate system performance in open-sky scenario; the first part of the UrbanNav-Deep dataset and the UrbanNav-Medium dataset are used to assess performance in sub-urban environments; while the second part of the UrbanNav-Deep dataset and the UrbanNav-Harsh dataset are used for evaluation in urban settings. 

\begin{figure}[htbp]
    \centering
    \subfigure[Open-Sky A]{\includegraphics[width=0.24\textwidth,height=0.12\textheight]{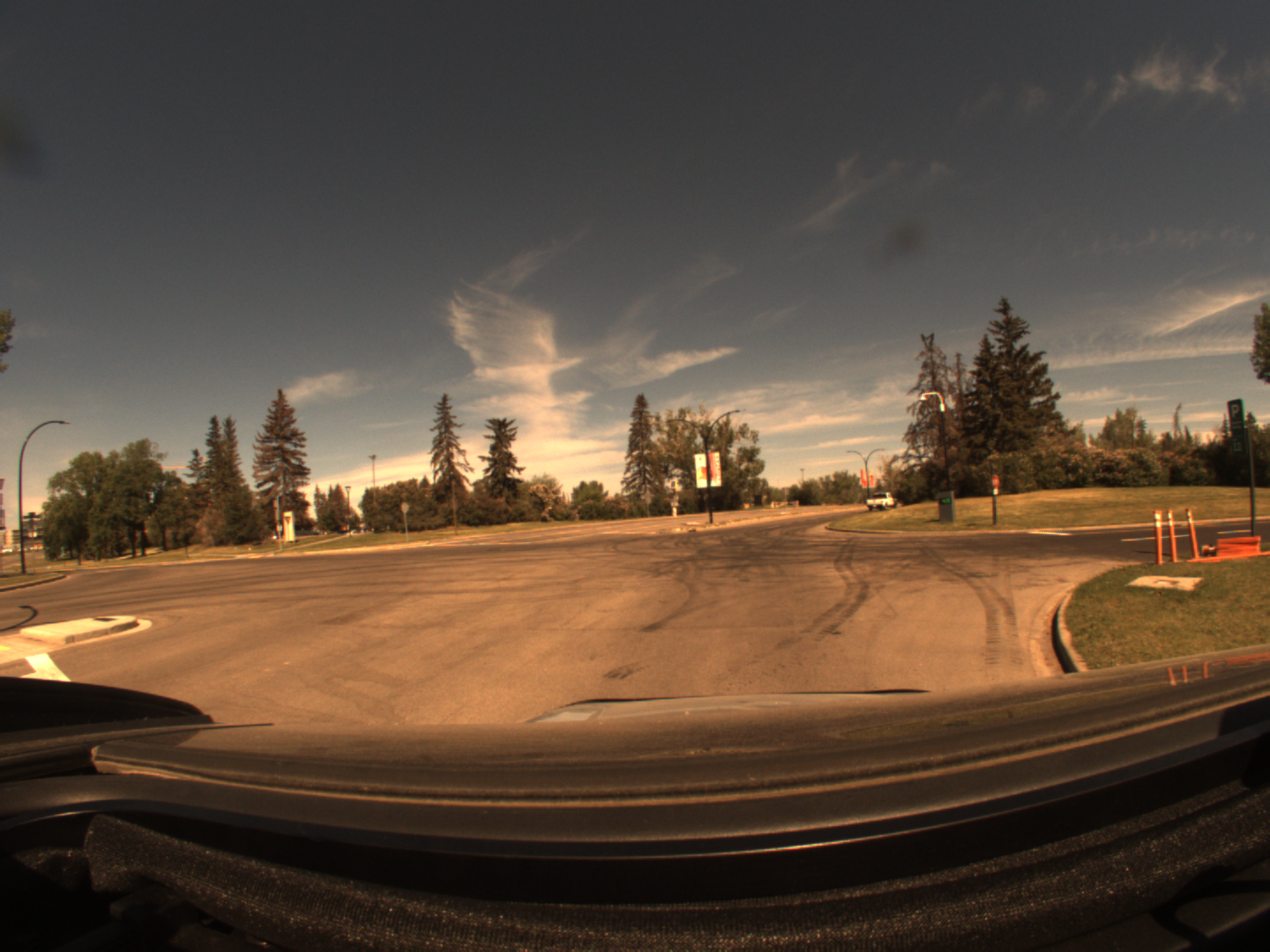}} \hfill
    \subfigure[Open-Sky B]{\includegraphics[width=0.24\textwidth,height=0.12\textheight]{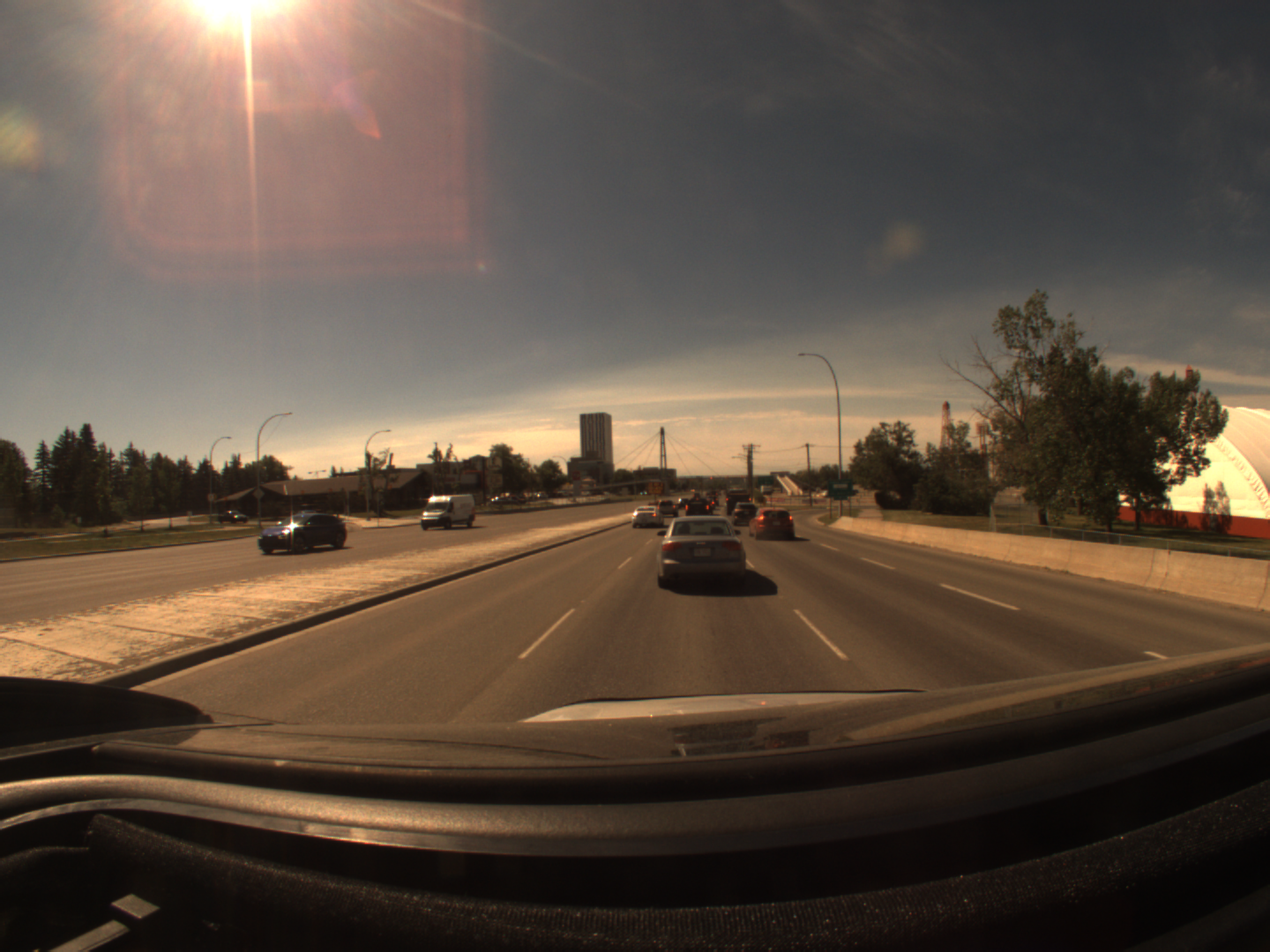}} \\
    \subfigure[Sub-urban A]{\includegraphics[width=0.24\textwidth,height=0.12\textheight]{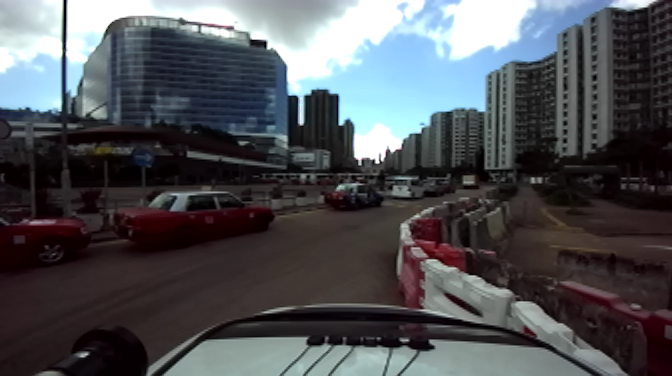}} \hfill
    \subfigure[Sub-urban B]{\includegraphics[width=0.24\textwidth,height=0.12\textheight]{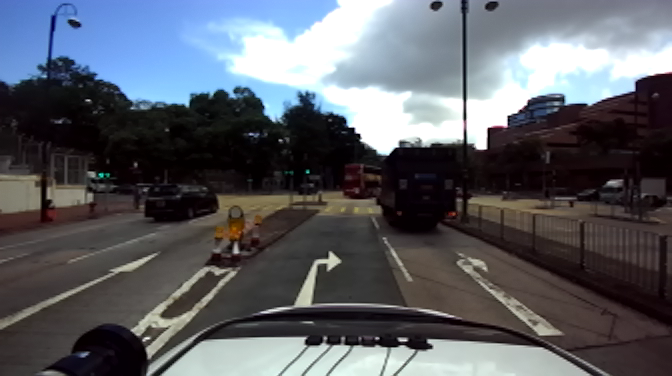}} \\
    \subfigure[Urban A]{\includegraphics[width=0.24\textwidth,height=0.12\textheight]{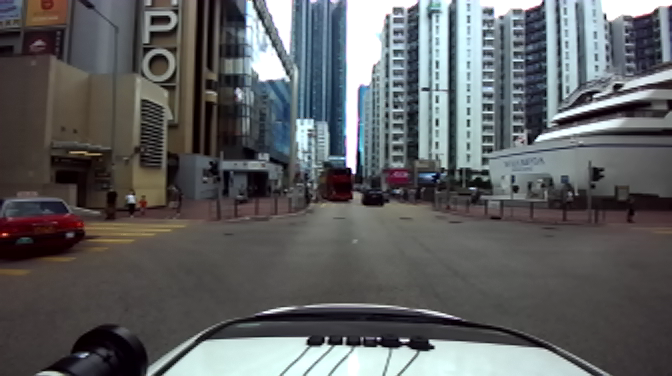}}\hfill
    \subfigure[Urban B]{\includegraphics[width=0.24\textwidth,height=0.12\textheight]{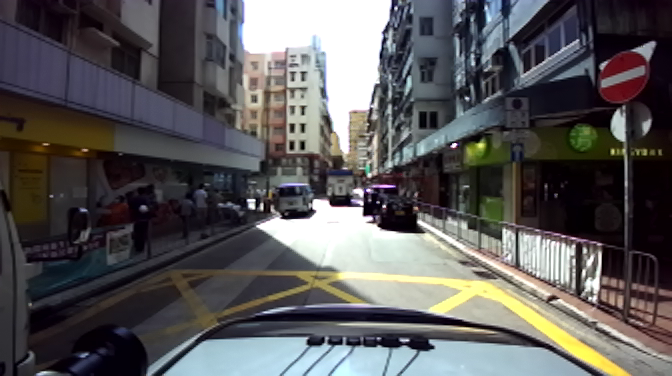}}
    \caption{Scenes from the validation datasets, featuring diverse environments.}
    \label{fig:2x2grid}
\end{figure}

Both Open-sky A and B datasets are collected near the University of Calgary, where A is captured near the CCIT building and B is primarily collected on a highway. The Sub-urban A dataset covers an area near Hung Hom in Hongkong, featuring side buildings, small tunnels, and loops. The Sub-urban B focus on the area near Tsim Sha Tsui Substation in the central area of Hongkong, encompassing medium-height buildings and numerous dynamic objects. The Urban A and B datasets are collected in deep urban environments where several GNSS signal blockages occur, resulting in frequent NLOS and multipath signals. Urban A is collected in the central area of Hung Hom, characterized by narrow roads surrounded by densely packed  building and heavy traffic. Urban B is collected in an ultra-dense urban canyon with narrow streets flanked by high-rising buildings, bridges, pedestrians and dense vehicle traffic.

Figure 5 illustrates the GNSS status in both sub-urban and urban environments. The available satellite number, Geometric Dilution of Precision (GDOP), and GNSS RTK fix status are visualized. To exclude GNSS outlier measurements, thresholds are applied: a minimum GNSS signal-to-noise ratio (SNR) of 30.0 $dB$ and a minimum satellite elevation of 7.0 degrees. When the GNSS solution is unavailable, the satellite number is plotted as zero. A fix status of 3 indicates a float ambiguity resolution for the GNSS RTK solution.

\begin{figure*}[htp]
  \centering
  \subfigure{\includegraphics[scale=0.43]{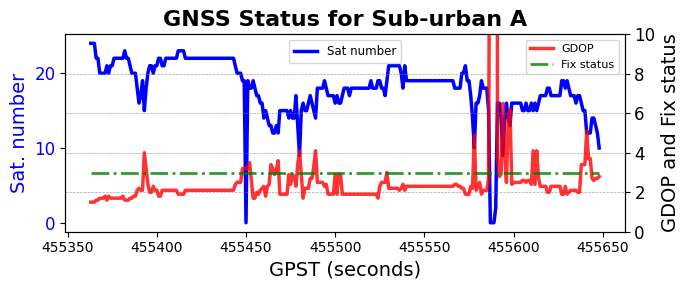}}\quad
  \subfigure{\includegraphics[scale=0.43]{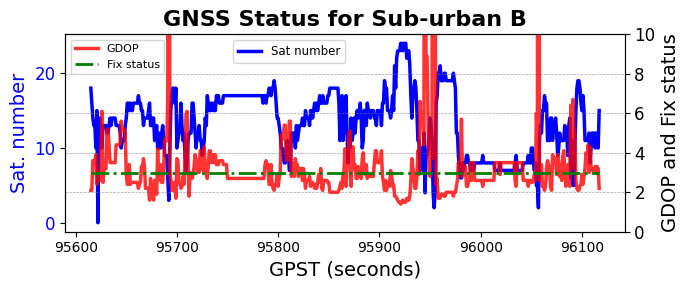}}\\
  \subfigure{\includegraphics[scale=0.43]{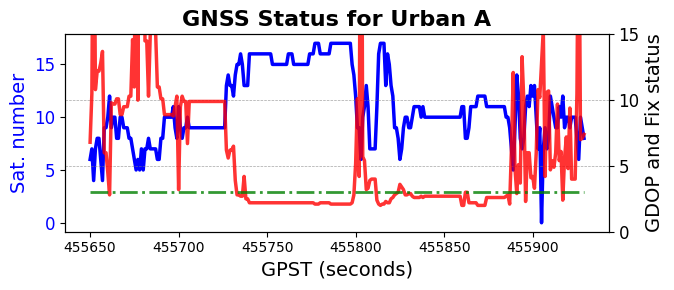}}\quad
  \subfigure{\includegraphics[scale=0.43]{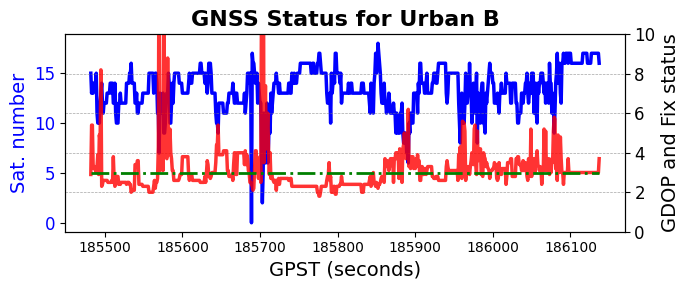}}
  \caption{Visualization of available satellite number, GDOP, and GNSS RTK fix status for Sub-urban and Urban scenarios.}
  \label{fig:gnss_status}
\end{figure*}

\subsection{Evaluation of Localization}

To validate the performance of GS-GVINS, we computed the APE of the navigation solutions from GS-GVINS against the ground truth and benchmark them against the tightly coupled solutions from GICI-LIB and IC-GVINS. All datasets are evaluated with successful GNSS/IMU and visual initializations. Table 1 summarizes the APE results for each environment type using the mentioned datasets. The values in bold indicate the lowest APE in each category. For Urban-B data tested with IC-GVINS, the solution is not applicable as the estimator diverges.

\begin{table}[ht!]
\centering
\renewcommand{\arraystretch}{1.5}
\captionsetup{font=small, justification=centering, singlelinecheck=false,labelsep=space, skip=5pt}
\caption{\\3D APE OF GS-GVINS, GICI, and IC-GVINS (METERS/DEGREES).}
\hspace{-0.2 cm}
\resizebox{0.5\textwidth}{!}{ 
\begin{tabular}{|p{1.5cm}|c|lll|} 
\hline
\multirow{2}{*}{\small\begin{tabular}[c]{@{}c@{}} \vspace{-0.2cm} Data\\ Environment\end{tabular}} 
    & \multicolumn{1}{l|}{\multirow{2}{*}{Data ID}} 
    & \multicolumn{3}{l|}{APE (Translation [m] / Rotation [$^\circ$])}                  \\ \cline{3-5} 
    & \multicolumn{1}{l|}{}                         
    & \multicolumn{1}{l|}{GS-GVINS} & \multicolumn{1}{l|}{GICI} & IC-GVINS \\ \hline
\multirow{2}{*}{\small Open-sky} & A 
    & \multicolumn{1}{l|}{0.073 / 3.748}         & \multicolumn{1}{l|}{\textbf{0.071} / \textbf{3.293}}     & 0.085 / 3.844      \\ \cline{2-5} 
    & B 
    & \multicolumn{1}{l|}{\textbf{0.084} / 3.154}         & \multicolumn{1}{l|}{0.081 / \textbf{3.028}}     & 0.266 / 3.623      \\ \hline
\multirow{2}{*}{\small Sub-urban} & A 
    & \multicolumn{1}{l|}{\textbf{1.873} / 3.622}         & \multicolumn{1}{l|}{1.952 / \textbf{3.562}}     & 6.644 / 5.562      \\ \cline{2-5} 
    & B 
    & \multicolumn{1}{l|}{\textbf{3.255 / 0.864}}         & \multicolumn{1}{l|}{3.934 / 1.012}     & 11.596 / 2.477      \\ \hline
\multirow{2}{*}{\small \ \ \  Urban} & A 
    & \multicolumn{1}{l|}{\textbf{4.521} / 1.646}         & \multicolumn{1}{l|}{4.584 / 1.647}     & 7.465 / \textbf{0.913}      \\ \cline{2-5} 
    & B 
    & \multicolumn{1}{l|}{\textbf{6.043} / 2.015}         & \multicolumn{1}{l|}{7.003 / \textbf{1.957}}     & \ \ \ \ - / -      \\ \hline
\end{tabular}}
\label{tab:example_table}
\end{table}

It is evident that GS-GVINS outperforms both GICI-LIB and IC-GVINS in the majority of the datasets. In GNSS-friendly Open-sky A and B environments, GS-GVINS and GICI demonstrate competitive performance. In these scenarios, GNSS RTK provides high-accuracy absolute positioning, which predominantly influences the solution estimation. Since the GNSS RTK factor implementations in GS-GVINS is adopted from GICI-LIB, the improvements in absolute pose estimation are limited. Additionally, as inertial and visual navigation have minimal influences in these scenarios, the following discussion will focus on the sub-urban and urban environments.

\begin{figure}[htbp]
\centering
{\includegraphics[width=1.0\columnwidth]{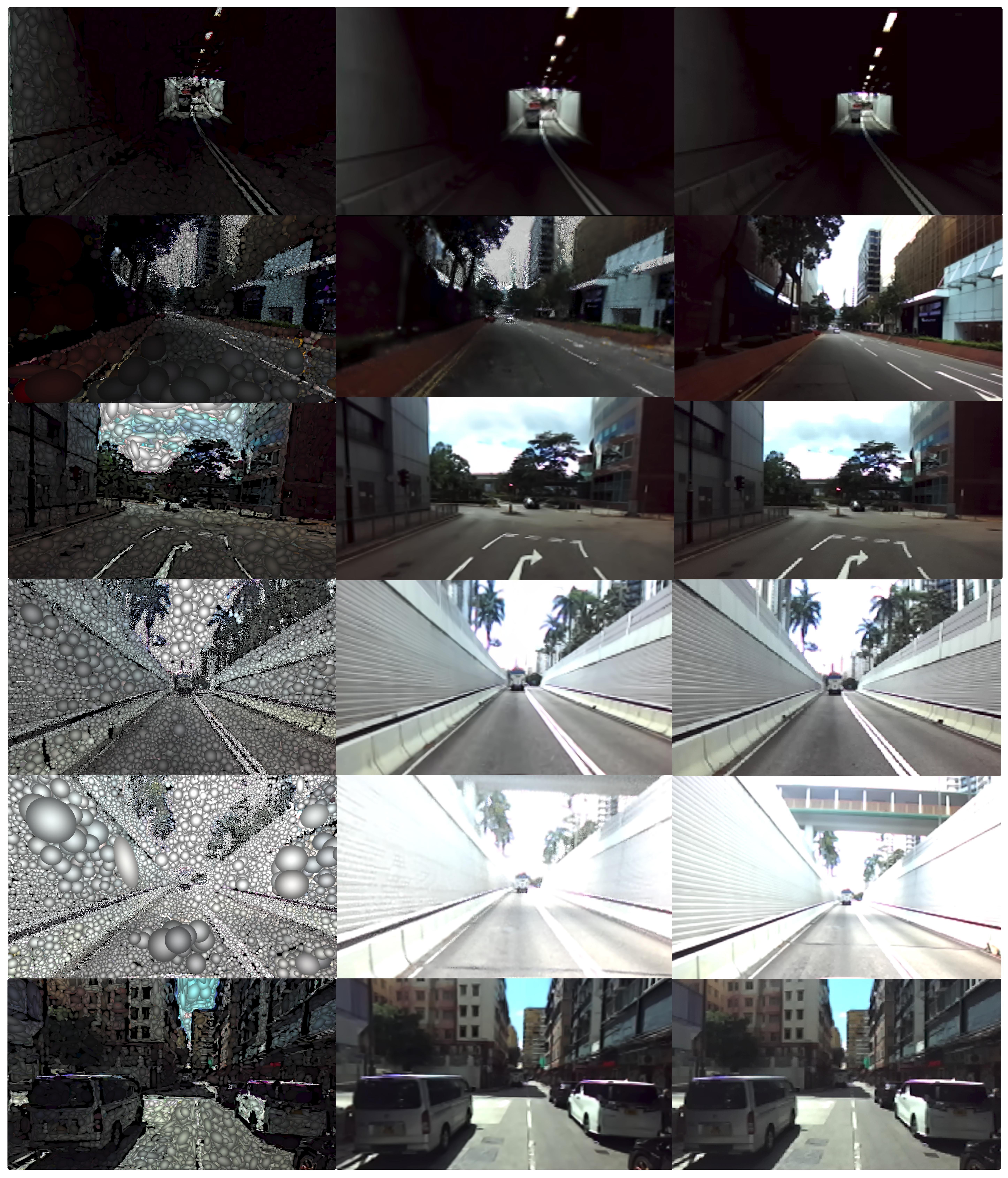}}
\caption{Visualization of 3DGS under different conditions. The left, middle and right column represents 3D Gaussians map, rendered image and groundtruth image, respectively.}
\end{figure}

\begin{figure}[htp]
  \centering
  \subfigure[Featureless environment]{\includegraphics[width= 4.2cm, height=2.55cm]{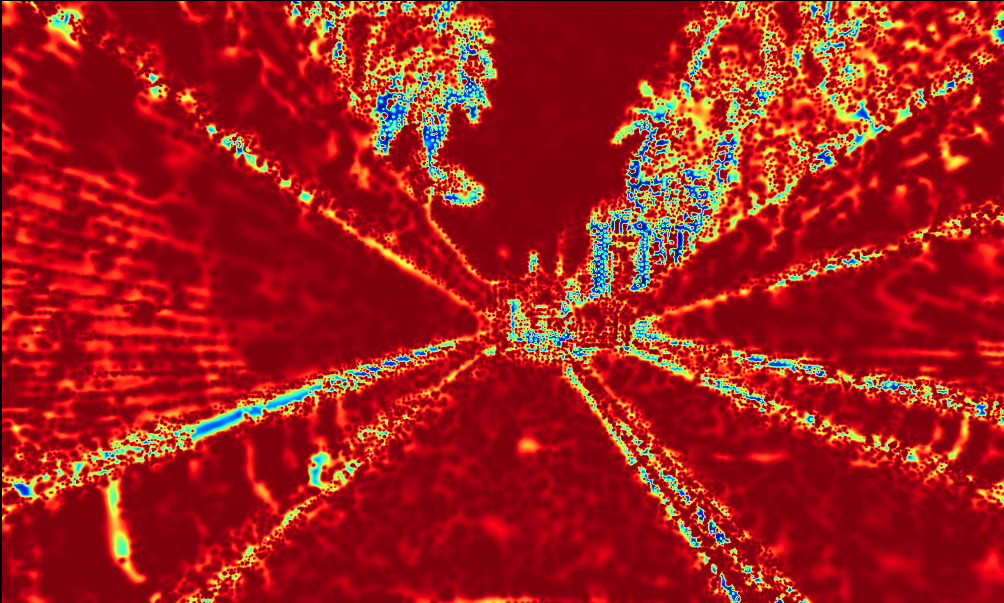}}\quad
  \subfigure[Static environment]{\includegraphics[width= 4.2cm, height=2.55cm]{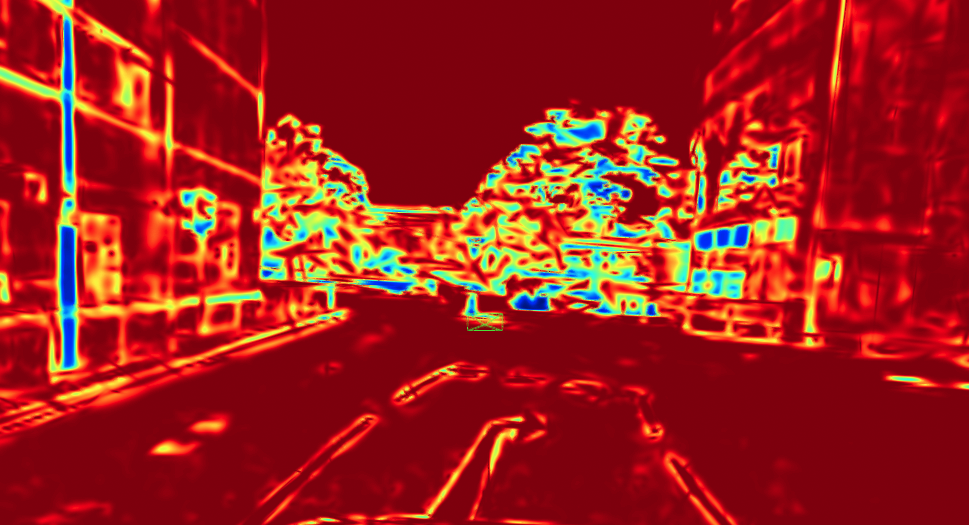}}\\
  \caption{Opacity shading: A jet colormap represents the optimized opacity of projected Gaussians.}
\end{figure}

\begin{figure}[htp]
  \centering
  \subfigure[3D Gaussians map]{\includegraphics[width= 4.2cm, height=2.55cm]{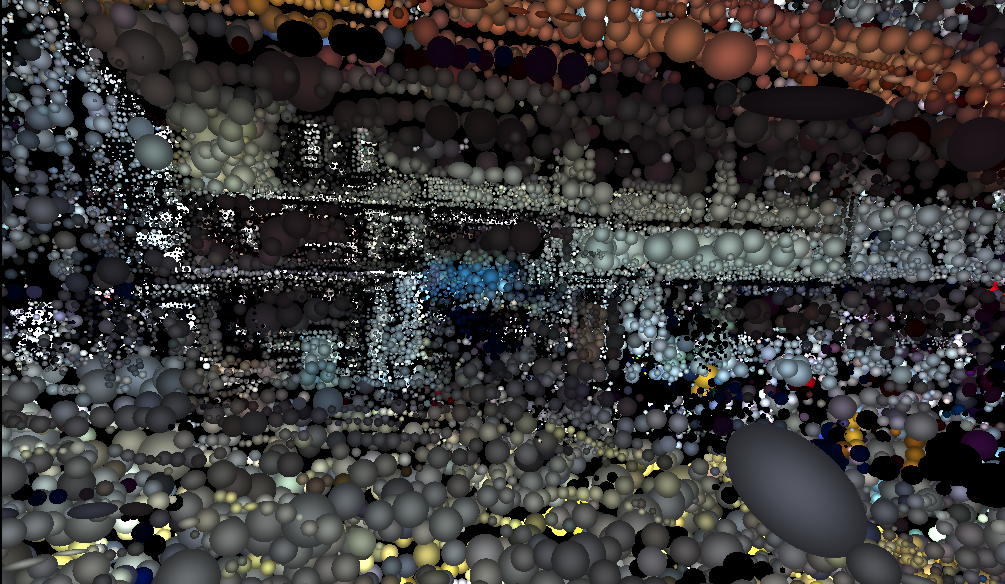}}\quad
  \subfigure[Rendered image]{\includegraphics[width= 4.2cm, height=2.55cm]{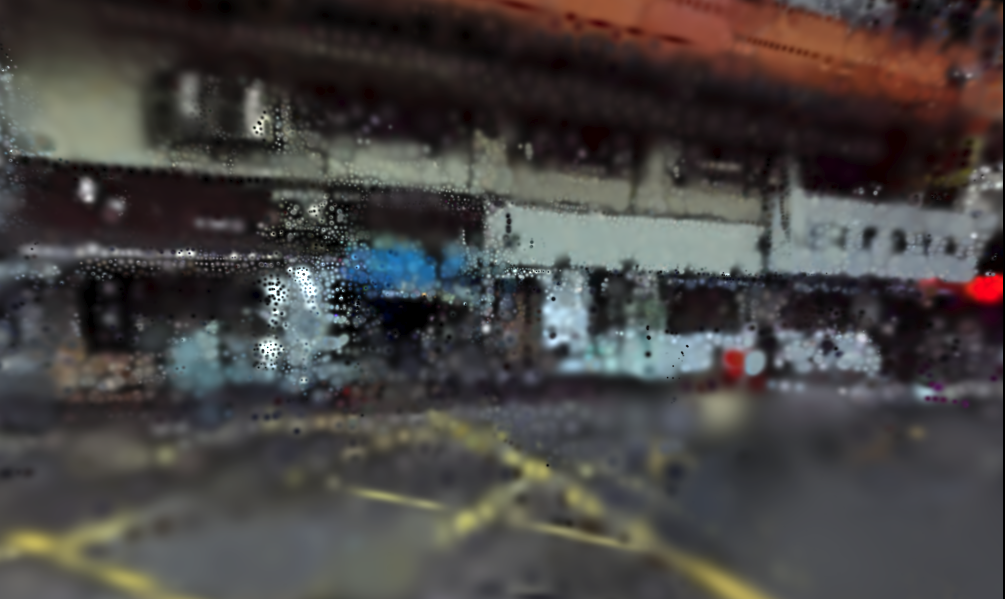}}\\
  \subfigure[Time shading]{\includegraphics[width= 4.2cm, height=2.55cm]{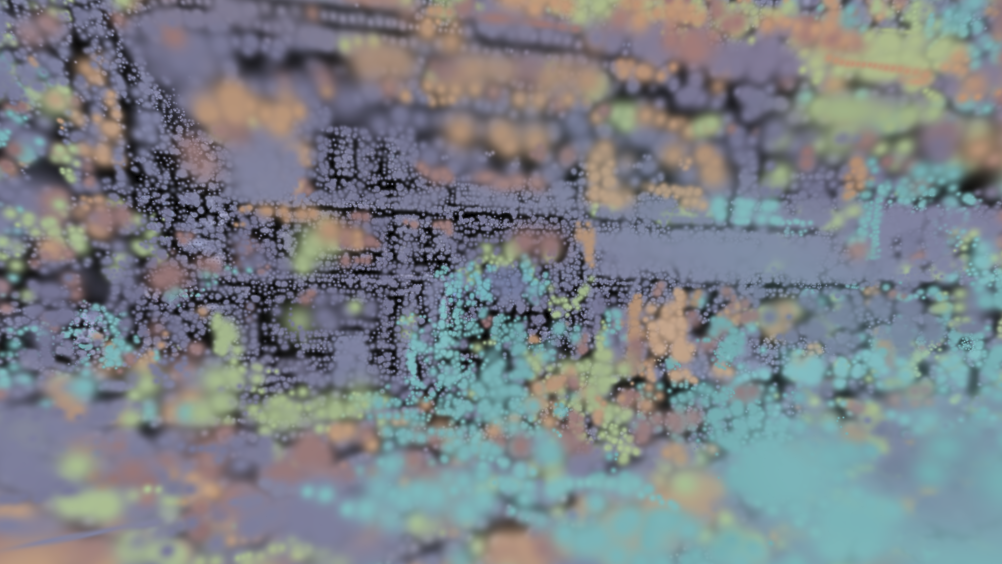}}
  \hspace{0.1cm}
  \subfigure[Ground truth]{\includegraphics[width= 4.2cm, height=2.55cm]{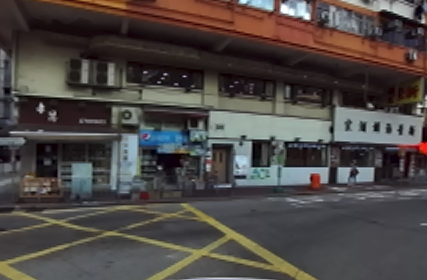}}
  \caption{3DGS in high dynamic condition}
\end{figure}

In both the sub-urban and urban datasets, GS-GVINS demonstrates significant improvements of 3D translation estimation over GICI and IC-GVINS. In Sub-urban A, GS-GVINS reduces 3D translation RMSE by 4.05 \% and 71.81 \%, compared to GICI and IC-GVINS, respectively. In Sub-urban B, GS-GVINS further reduces translation errors by 17.26 \% (GICI) and 71.93 \% (IC-GVINS), along with a 14.62 \% reduction in 3D rotation RMSE. In the more challenging environment, Urban A, GS-GVINS improves translation accuracy compared to GICI and IC-GVINS by 1.37 \% and 39.44 \%, respectively. Despite GS-GVINS achieves a slightly better rotation estimation than GICI, the lowest 3D rotation RMSE (0.913 degrees) is achieved by IC-GVINS. In Urban B, GS-GVINS achieves a 13.71 \% reduction in 3D translation errors compared to GICI, further demonstrating its robustness in translation estimation under challenging conditions.

Despite of the improvements of 3D translation estimation by GS-GVINS, its improvements in 3D rotation estimation remain limited. This limitation in rotation accuracy is partly due to the depth initialization strategy for the monocular setting. To compromise the complex and dynamic driving environments, we uniformly initialize the depth of each new Gaussian based on the approximated median depth of the scene with added noise, then refine it during rasterization. However, this coarse depth initialization, combined with the lack of multi-view optimization for 3D Gaussians, can introduce rotation scale errors, which limit the accuracy of 3D rotation estimation in GS-GVINS.

\begin{figure*}[htp]
  \centering
  \subfigure[Sub-urban A]{\includegraphics[width= 7.2cm, height=4.9cm]{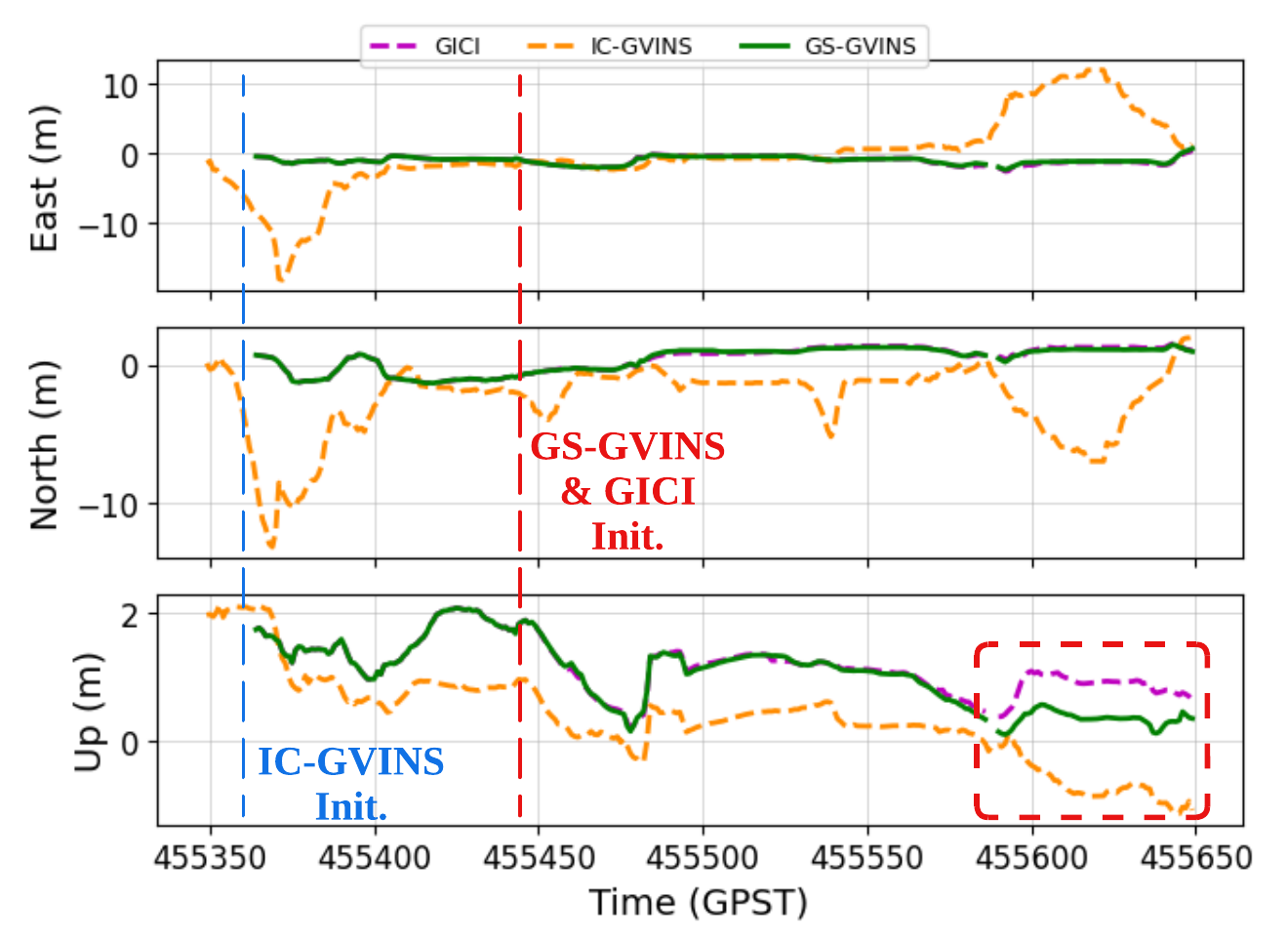}}\quad
  \subfigure[Sub-urban B]{\includegraphics[width= 7.3cm, height=4.9cm]{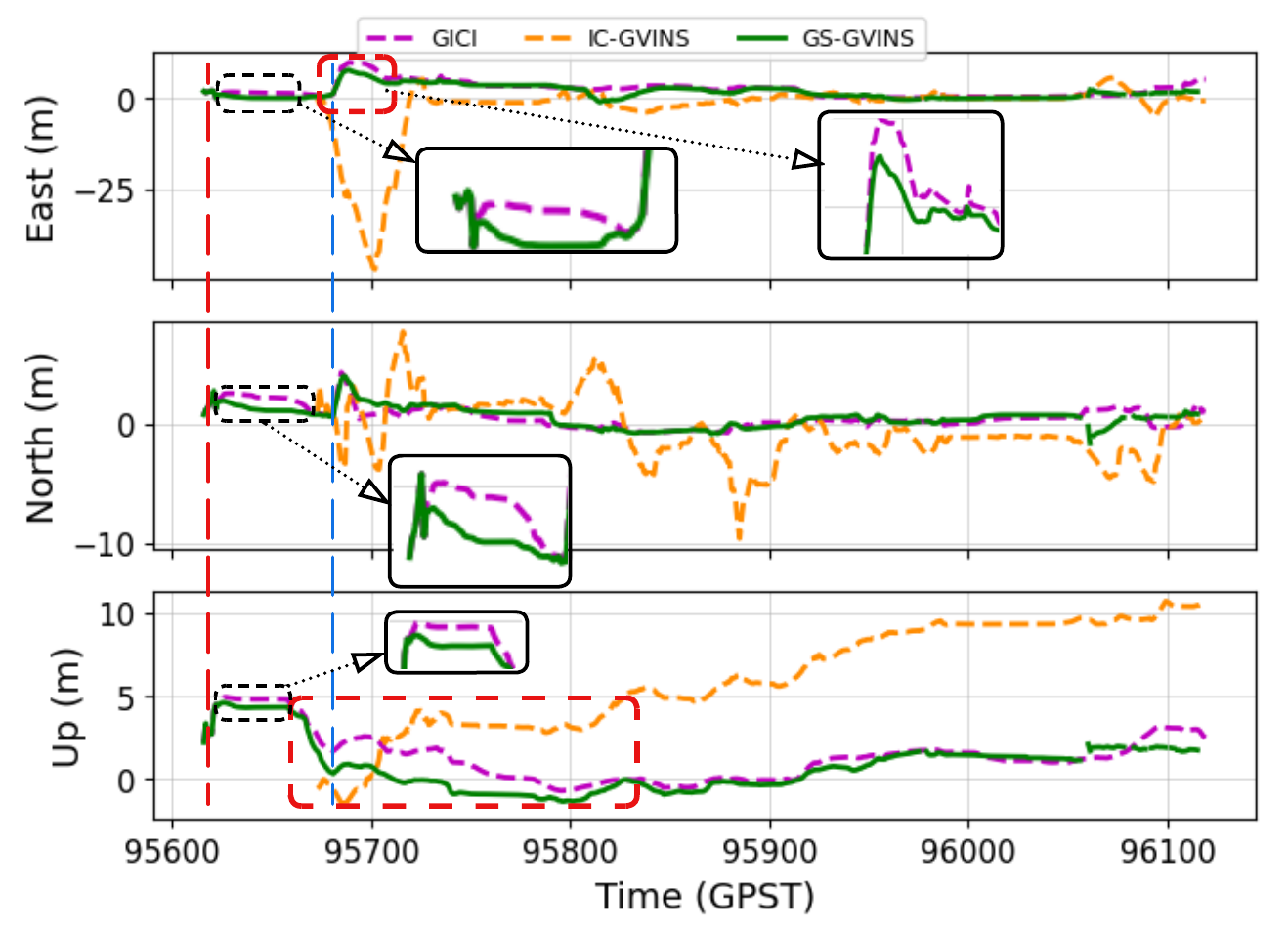}}

  \vspace{-0.5em}
  
  \subfigure[Urban A]{\includegraphics[width= 7.2cm, height=4.9cm]{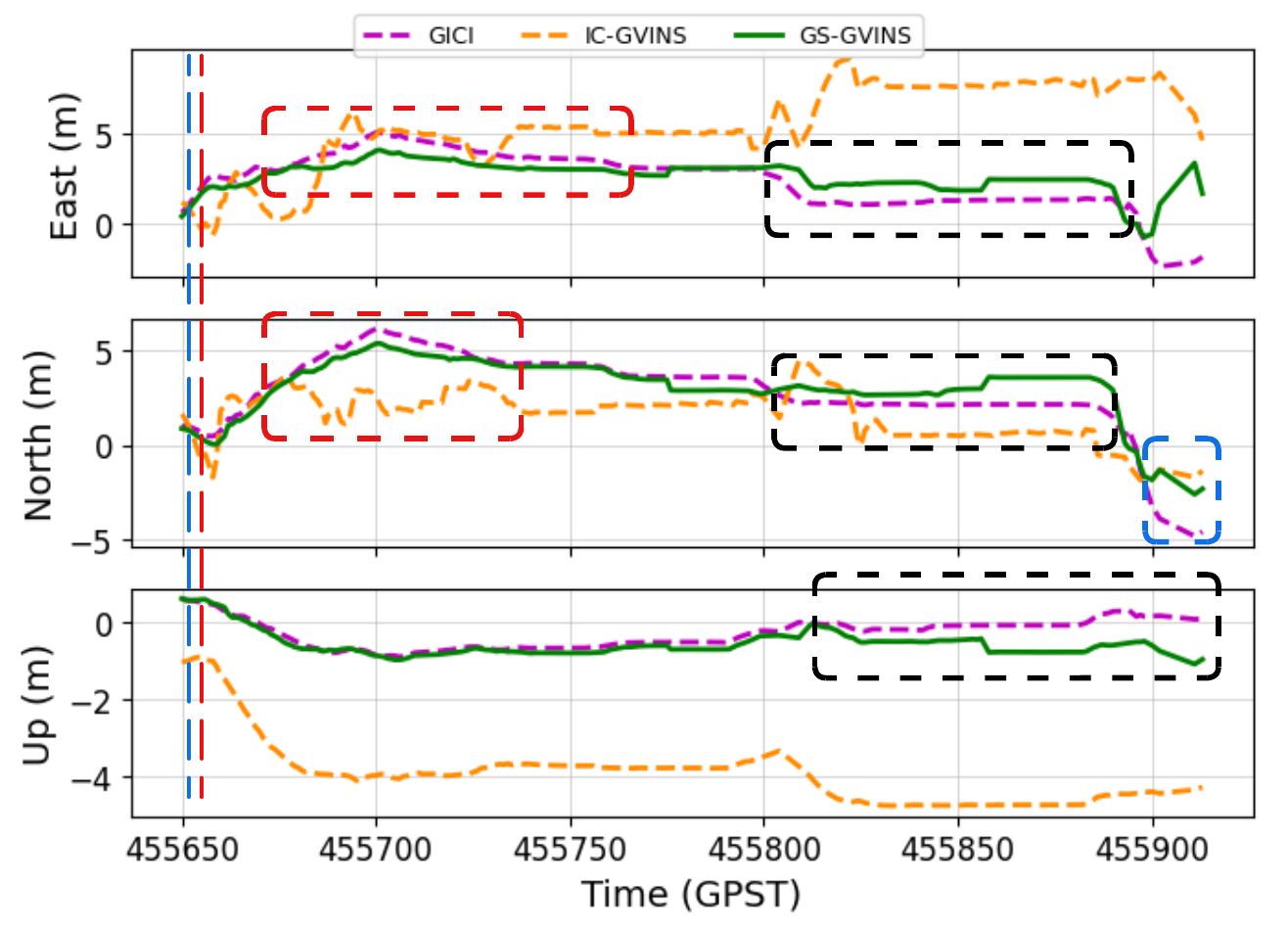}}
  \subfigure[Urban B]{\includegraphics[width= 7.2cm, height=5.0cm]{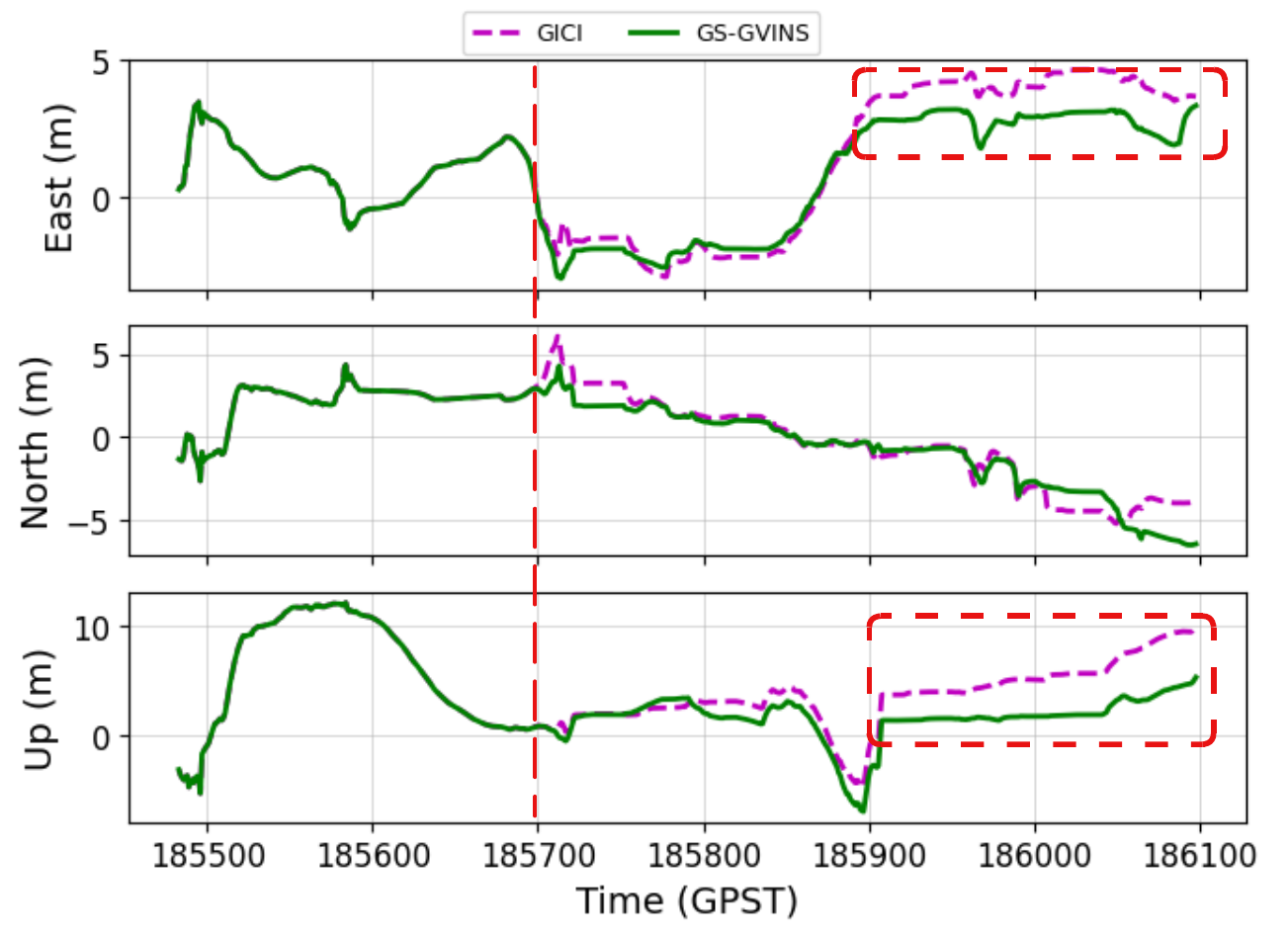}}
  \caption{3D Absolute Translation Error in ENU directions over GPS time.}
\end{figure*}

 \begin{figure}[htp]
  \centering
  {\includegraphics[width= 4.0cm, height=2.8cm]{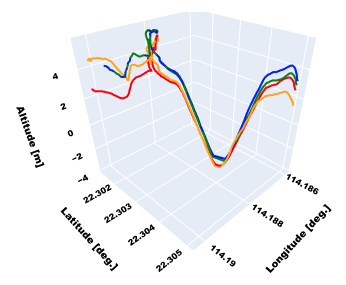}}\quad
  {\includegraphics[width= 4.0cm, height=2.9cm]{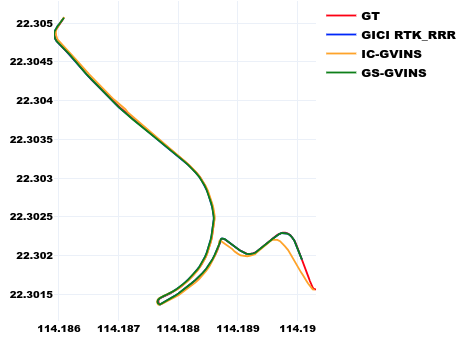}}\\
  {\includegraphics[width= 3.7cm, height=2.6cm]{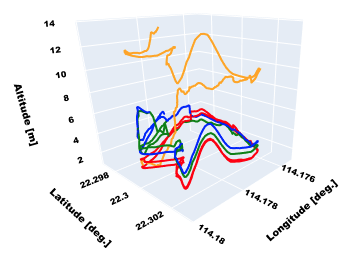}}
  {\includegraphics[width= 4.0cm, height=2.8cm]{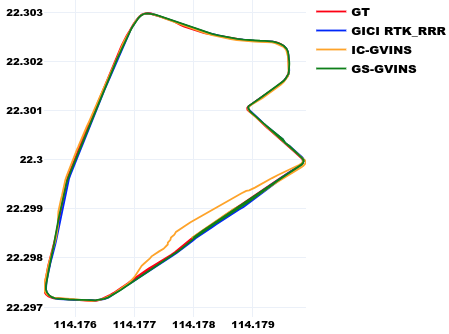}}\\
  {\includegraphics[width= 3.8cm, height=2.8cm]{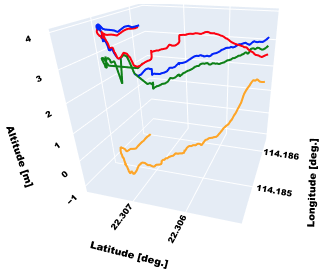}}
  {\includegraphics[width= 4.1cm, height=3.0cm]{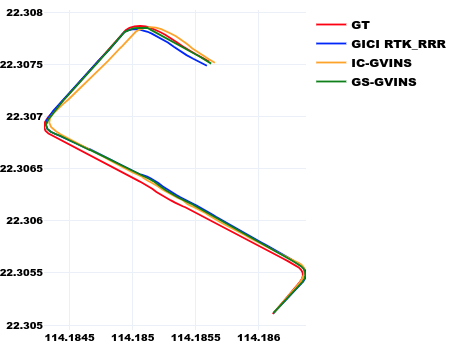}}\\
  {\includegraphics[width= 3.7cm, height=2.8cm]{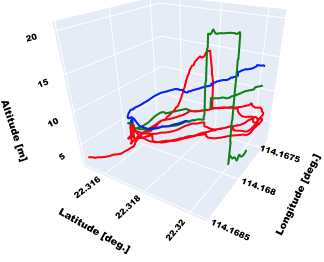}}
  {\includegraphics[width= 4.1cm, height=2.9cm]{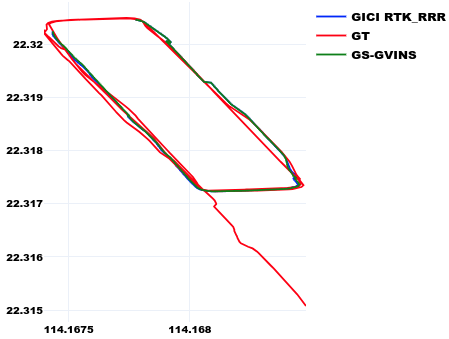}}
  \caption{3D and 2D trajectory plots for Sub-urban A, Sub-urban B, Urban A and Urban B (from top to bottom).}
\end{figure}

Figure 9 illustrates the 3D absolute translation error over GPS time for each dataset in the ENU directions. The red dashed vertical line represents the initialization time for GS-GVINS and GICI, as GS-GVINS adopts its initialization algorithm from GICI, resulting in identical completion times. The 3D Gaussian map is initialized in the next image frame after feature-based visual initialization. The blue vertical line marks the initialization time for IC-GVINS. Due to the different initialization strategies of GICI and IC-GVINS, and the dependency of IC-GVINS on the availability of GNSS RTK solutions, their initialization times vary. In Figure 9 (d), the absolute translation error of IC-GVINS is not included, as the estimator diverges.  

In Figure 9 (a), a notable reduction in the vertical (up) translation error from GS-GVINS is observed indicated by the red dashed box. This improvement can also be observed near the right-side tail of the 3D trajectory in Figure 10. The corresponding driving scenario is depicted in the first, fourth and fifth rows of Figure 6, where the vehicle is going through an underground tunnel. Extreme camera exposure conditions are observed during this time period, where the tracked visual features can be lost or mismatched. However, 3DGS maintains high-fidelity rendering despite these conditions. Furthermore, the fourth row of Figure 6 highlights a featureless scene, which makes visual feature detection and tracking challenging. Despite this, 3DGS accurately renders the scene while preserving fine details. Figure 7 (a) displays the rendering from the opacity shader, where the opacity values extracted from the Gaussian rendering are visualized using a jet colormap (warmer colors indicate higher opacity values, and colder colors indicate lower opacity values). The comb-like patterns on the tunnel walls are precisely textured by the optimized opacity values. Although GNSS performance deteriorates during this period, as reflected in Figure 5, the 3DGS factor effectively bounds the vertical translation errors.

In Figure 9 (b), GS-GVINS significantly reduces translation errors in the ENU directions compared to GICI between 95610s and 95660s (black box). This period coincides with frequent fluctuations in GDOP values and a sharp decline in satellite numbers, indicating poor GNSS performance. The corresponding scenario is shown in the third row of Figure 6, where the vehicle is reaching to a stop. The motion-aware pruning mechanism aids in achieving accurate rendering in this case. Figure 7 (b) presents the opacity map, revealing a well-defined outline and structure of objects in the scene, which helps establish better zero-motion constraints from the 3DGS factor. Around 95680s, a sharp increase in GDOP value, combined with fewer than five available GNSS satellites, results in large error spikes in both east and north directions. However, GS-GVINS significantly reduces the east and vertical positioning error (red box) compared to both GICI and IC-GVINS, bringing the vertical positioning error close to zero. The second row of Figure 6 illustrates this environment, where dense tree canopies and advertising boards cast shadows over the left side of the street, making feature tracking difficult.

Figures 9 (c) and (d) display ENU error plots for more urbanized environments (Urban A and B). In Urban A, prolonged low satellite availability and extreme GDOP values (indicated by the red box) force the estimator to rely more on inertial and visual navigation. The last row of Figure 6 characterizes the street scene, featuring narrow roads, high-rise buildings, and parked cars, all of which are mapped by 3DGS. In this scenario, the 3DGS factor effectively constraints the growth of translation errors, as GNSS service is largely unavailable. As shown in Figure 9 (c), GS-GVINS improves both east and north translation errors to GICI and IC-GVINS as show in the red box. However, during the periods indicated by the black boxes, 3DGS factor seems to hinder translation error reduction. As the DOP values improve and the number of available satellites increases during this period, the influence of the 3DGS factor likely diminishes, weakening the impact of GNSS-based positioning during optimization. GS-GVINS achieves a significant reduction in north-direction error compared to GICI, as highlighted by the blue box, improving vehicle turning performance as shown in the third row of Figure 10 (near the end of the trajectory). Most of the cases discussed earlier also occur in Urban B, where GS-GVINS constrains the east and up error highlighted by the red box in Figure 9 (d). During these periods, with varying illumination caused by urban shadows and the presence of numerous pedestrians, the 3DGS factor effectively controls error drift in feature-based visual-inertial navigation.

\section{Conclusion}

In this paper, we proposed the GS-GVINS, a tightly-integrated GNSS-Visual-Inertial Navigation System augmented by 3DGS, incorporating a mapping loss-based weighting scheme for 3DGS factor and a motion-aware Gaussian pruning mechanism. Comprehensive evaluation of pose estimation have been conducted using the data from complex, large-scale outdoor environments. Experimental results demonstrate significant improvements, particularly in translation accuracy, compared to SOTA sensor-fusion libraries GICI and IC-GVINS. 

Additionally, the results highlight the robustness of 3DGS in handling challenging conditions such as extreme exposure, featureless scenes, and shadowed areas—scenarios that typically degrade feature-tracking-based navigation. Furthermore, the pruning mechanism enhances mapping quality during extreme vehicle dynamics, improving pose estimation accuracy in both low- and high-dynamic motion.

Since GS-GVINS operates in a pseudo-real-time configuration, we believe it can achieve real-time performance with sufficient computational resources. Moreover, GS-GVINS has the potential for high-fidelity map reconstruction in typical driving scenarios, particularly when supplemented with accurate depth information from external sensors. A well-reconstructed 3D Gaussian map can further enhance GNSS performance, especially by improving outlier detection, where GNSS signal rays are traced against the 3D reconstructed Gaussian map to identify NLOS signals or multipath.

\section*{Acknowledgments}
We acknowledge the support of the Natural Sciences and Engineering Research Council of Canada (NSERC); we also extend our appreciation to Cheng Chi for providing support and discussion regarding the GICI library.

\end{document}